\documentclass{article}
\usepackage{etoolbox}       %
\usepackage{tabularx}       %

\usepackage{graphicx}  %

\usepackage[parfill]{parskip}
\usepackage{amsmath}
\usepackage{amsthm}
\usepackage{mathtools}  %
\usepackage[dvipsnames]{xcolor}         %
\usepackage{float}  %
\usepackage{listings}  %

\newtoggle{arxiv}
\toggletrue{arxiv}

  \usepackage[parfill]{parskip}
  \usepackage[citestyle=authoryear-comp,sorting=nyt,maxbibnames=99,backend=biber]{biblatex}
  \newcommand{\citep}{\parencite}
  \newcommand{\citet}{\textcite}
  \newlength{\defbaselineskip}
  \RequirePackage[T1]{fontenc}
  \RequirePackage[tt=false, type1=true]{libertine}
  \RequirePackage[varqu]{zi4}
  \RequirePackage[libertine]{newtxmath}

\usepackage{bm}

\usepackage[inline]{enumitem}
\usepackage{booktabs}
\usepackage[dvipsnames]{xcolor}  %
\usepackage{subcaption}
\usepackage{multirow}
\usepackage{wrapfig}
\usepackage{algorithm,algorithmicx,algpseudocode}
\usepackage{nicematrix}

\usepackage[frozencache,cachedir=.]{minted}

\usepackage[colorlinks=true,linkcolor=blue,citecolor=magenta,urlcolor=red]{hyperref}
\usepackage[capitalise,noabbrev]{cleveref}

\usepackage{import}

\usepackage{pifont}

\newcommand*\samethanks[1][\value{footnote}]{\footnotemark[#1]}

  \title{Wonderful Matrices: More Efficient and Effective Architecture for Language Modeling Tasks}
  \usepackage{authblk}
  \author[$^1$]{Jingze Shi\thanks{Algorithm Design and Experiment Verification.}}
  \author[$^2$]{Bingheng Wu\thanks{Feature Analysis and Datasets Processing.}}
  \author[$^3$]{Lu He\samethanks[2]}
  \author[$^4$]{Luchang Jiang\samethanks[2]}
  \affil[ ]{Independent Researcher}
  \affil[ ]{{\texttt{losercheems@gmail.com}}}
  \affil[ ]{{\texttt{wubingheng52136@gmail.com}, \texttt{2196187034@qq.com}, \texttt{auroral.sunflower@gmail.com}}}
  \date{}

\begin{document}

  \maketitle

\begin{abstract}
% 我们证明了内积形式的位置编码在状态空间对偶算法的可用性, 并研究了不同位置嵌入在混合二次因果自注意力与状态空间对偶算法的有效性. 我们提出了内函数注意力与动态掩码, 可以提高注意力算法的表达性并避免序列噪声大幅影响关注分数的准确性. 我们还设计了交叉领域混合专家, 可以提高稀疏激活的前馈网络的细粒度, 而保持参数利用与检索的高效性. 这些方法的结合构成了我们的基础模型架构: Wonderful Matrices. 我们在语言建模任务上进行了实验, 发现 Wonderful Matrices 在处理复杂语言任务时更具高效性和有效性.
We prove the availability of inner product form position encoding in the state space duality algorithm and study the effectiveness of different position embeddings in the hybrid quadratic causal self-attention and state space duality algorithms. We propose inner function attention with dynamic mask, which can improve the expressiveness of the attention algorithm and avoid the sequence noise significantly affecting the accuracy of the attention score. We also design cross domain mixture of experts, which can improve the granularity of the sparse activation feedforward network while maintaining the efficiency of parameter utilization and retrieval. The combination of these methods constitutes our foundation model architecture: Wonderful Matrices. We conduct experiments on the language modeling task and find that Wonderful Matrices are more efficient and effective in handling complex language tasks.
\end{abstract}

\section{Introduction}
\label{sec:introduction}

% 高效的算法旨在有限的状态中压缩信息, 使其能够在有限的状态空间中存储尽可能多的有用信息, 而有效的算法旨在存储所有信息状态, 并构建信息之间的依赖关系, 以避免捕获偏差信息.
Efficient algorithms aim to compress information in a limited state so that it can store as much useful information as possible in a limited state space, while effective algorithms aim to store all information states and build dependencies between information to avoid capturing biased information.

% Transformers 架构在现代深度学习语言建模中很受欢迎, 它可以直接捕获序列中任意两个元素之间的关系, 并有效处理长距离依赖问题. 然而, 该架构有两个主要缺点. 首先, 在处理长序列时, 其自注意力算法的二次复杂度和缓存大小限制了处理长上下文的能力. 其次, Transformers 缺乏单个汇总状态, 这意味着每个生成的令牌必须在整个上下文上计算.
Transformers~\citep{vaswani2017attention} Architecture is popular in modern deep learning language modeling, which can directly capture the relationship between any two elements in the sequence and effectively deal with long-distance dependency problems. However, the architecture has two main drawbacks. First, when dealing with long sequences, the quadratic complexity of its causal self-attention algorithm and the cache size limit the ability to process long contexts. Second, Transformers lack a single summary state, which means that each generated token must be computed over the entire context.

% 与此同时, 选择性状态空间模型应运而生. 选择性状态空间模型通过其选择性状态更新算法存储有效的相关信息, 平衡相关矩阵的二次和线性计算方法, 在训练期间实现序列长度的线性缩放, 并在生成期间保持恒定的状态大小. 此外, 由于其线性递归状态更新机制, 选择性状态空间模型具有单个汇总状态. 然而, 选择性状态空间模型也有一个主要缺点: 其状态不随序列长度扩展, 信息压缩不可避免地导致信息丢失.
Concurrently with this, the State Space Model (Mamba2~\citep{mamba2}) came into being. Mamba2 stores effective relevant information through its selective state update algorithm, balances the quadratic and linear calculation methods (state space duality) of relevant matrices, achieves linear scaling of sequence length during training, and maintains a constant state size during generation. In addition, due to its linear recursive state update mechanism, Mamba2 has a single summary state. However, Mamba2 also has a major drawback: its state does not expand with the sequence length, and information compression inevitably leads to information loss.

% 为了构建既高效又有效的模型, 关键是平衡压缩信息状态和存储所有信息状态之间的关系. 我们的主要目标是将状态空间对偶算法与二次因果自注意力算法相结合, 以克服它们各自的局限性. 这种混合算法基础模型架构虽然会损失一部分单个算法在特定任务上的极致优秀性, 但是将会具有过滤信息、长上下文中的长期依赖关系、汇总状态、高效学习和低内存使用的综合能力. 本文旨在进一步探讨如何将选择性状态空间算法与二次自注意力算法相结合, 与具有交叉领域通用知识的混合专家相配合, 来构建一个比 Transformers 或 Mamba 更具综合能力的基础模型架构.
To build a model that is both efficient and effective, the key is to balance the relationship between compressing information states and storing all information states. Our main goal is to integrate the state space duality(referred to as SSD in the following) algorithm with the quadratic causal self-attention(referred to as QCAttn in the following) algorithm to overcome their respective limitations. Although this hybrid algorithm foundation model architecture will lose some of the extreme excellence of a single algorithm in a specific task, it will have the comprehensive ability of information filtering, long-term dependencies in long contexts, summary states, efficient learning, and low memory usage. This paper aims to further explore how to combine the selective state space algorithm with the quadratic self-attention algorithm, and cooperate with the mixture of experts with cross-domain general knowledge to build a foundation model architecture that is more comprehensive than Transformers or Mamba.

\begin{figure}[t]
    \centering
    \includegraphics[width=\linewidth]{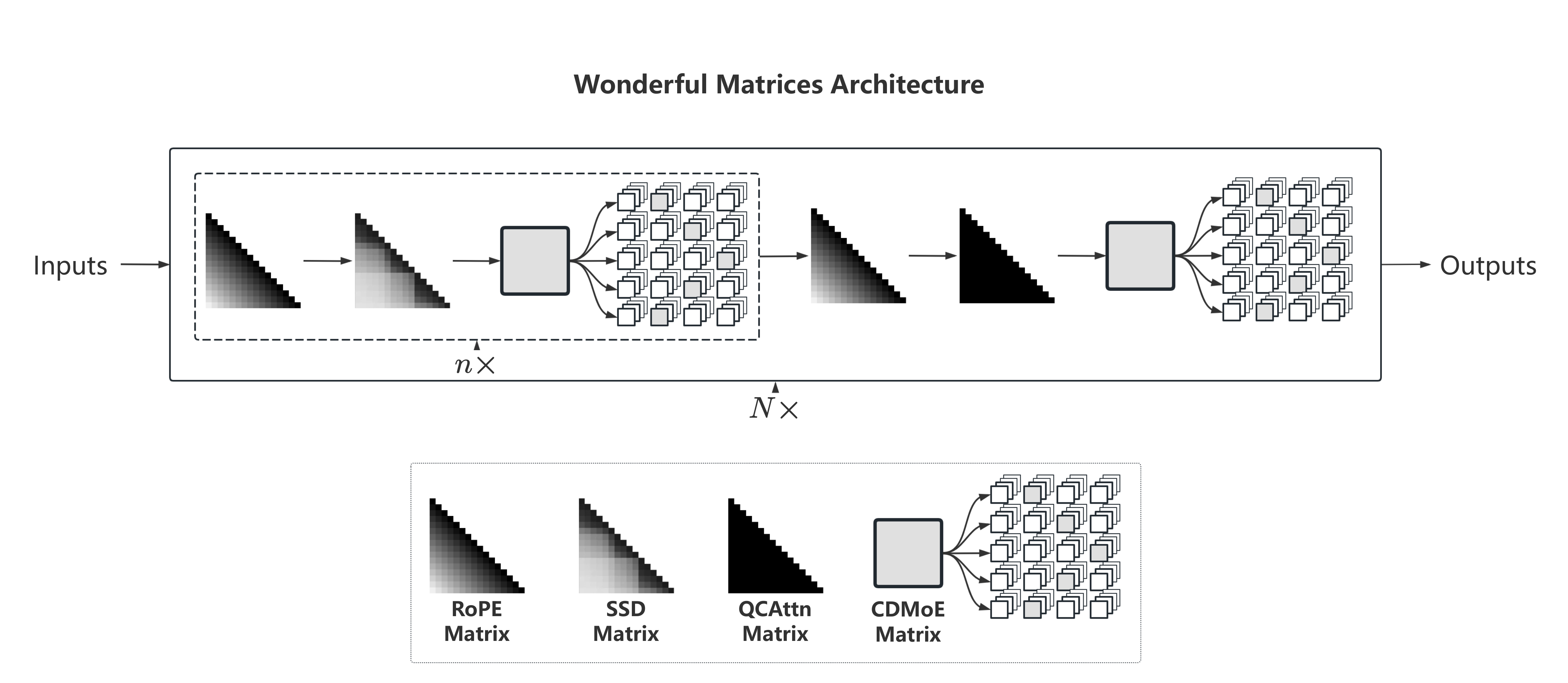}
    \caption{
      \textbf{Wonderful Matrices Architecture}.
      % 展示了 Wonderful Matrices 架构中使用的矩阵, 包括旋转位置编码矩阵, 状态空间对偶矩阵, 二次因果自注意力矩阵, 交叉领域混合专家矩阵, 以及使用这些矩阵的过程. 这些矩阵的具体结构和算法将在后续章节中详细介绍.
      Shows the matrices used in the Wonderful Matrices architecture, including the rotary position embedding matrix, state space duality matrix, quadratic causal self-attention matrix, cross domain mixture of experts matrix, and the process of using these matrices.
      The specific structure and algorithm of these matrices will be detailed in subsequent chapters.
    }
    \label{fig:wonderful_matrices}
\end{figure}

% 结合 SSD 算法和 QCAttn 算法的关键是有效整合位置信息. 在 Mamba~\citep{gu2023mamba} 中, 位置信息是通过因果卷积隐式提供的, 并使用矩阵 D 跳过选择性状态空间算法的输入和输出, 重新延续离散位置信息. 在 Mamba2~\citep{mamba2} 中, 累积乘积被提出, 允许两个位置相互作用, 这是一种相对位置嵌入的形式. 然而为了位置编码而进行额外的卷积运算较为费时, 而递归式的位置编码仅适用于 SSD 算法, 无法作用于 QCAttn. 我们需要一个统一的位置编码来保证序列的位置信息在两种算法中的一致性. 我们证明了旋转位置编码在 SSD 算法中的可用性, 以统一位置编码的方式, 保证了两种算法中的位置信息的一致性.
\paragraph{Position Encoding.}
The key to combining the SSD algorithm with the QCAttn algorithm is the effective integration of positional information. In Mamba~\citep{gu2023mamba}, the position information is implicitly provided by causal convolution, and matrix D is used to skip the connection to the input and output of the selective state space algorithm, re-continuing the discrete positional information. In Mamba2~\citep{mamba2}, the cumulative product is proposed to allow two positions to interact, which is a form of relative positional embedding. However, additional convolution operations for position encoding are time-consuming, and recursive position encoding is only applicable to the SSD algorithm and cannot be applied to QCAttn. We need a unified position encoding to ensure the consistency of positional information in the two algorithms. We have proven the availability of rotary position embedding in the SSD algorithm to ensure the consistency of positional information in the two algorithms.

% 提升序列变换的高效性需要减少 QCAttn 在整个序列变换层所占的比例, 而为了保持序列变换的有效性, 我们需要增加 QCAttn 的表达能力. grouped-query attention发现, 通过将多个查询作为一组来对应一个键值, 相较于多头注意力仅有略微的质量衰减, 而大幅减少加载键值的内存带宽. 那么相反, 增加值的表达性则会提升建模质量, 然而我们不能简单的增加值的数量, 牺牲效率来换取效果, 这与GQA的初衷相悖. 我们提出了内函数注意力与动态注意力掩码方法, 将 QCAttn 中的 Value 从对输入的线性映射修改为一个可以存储更多信息的启发式函数, 并防止过多信息噪声影响注意力分数矩阵的计算.
\paragraph{Algorithm Mixing.}
Improving the efficiency of sequence transformation requires reducing the proportion of QCAttn in the entire sequence transformation layer, and to maintain the effectiveness of sequence transformation, we need to increase the expressive power of QCAttn. Grouped-query attention~\citep{ainslie2023gqa} found that by corresponding multiple queries to a group of keys and values, there is only a slight quality degradation compared to multi-head attention, but a significant reduction in memory bandwidth for loading keys and values. Conversely, increasing the expressiveness of values will improve the quality of modeling, but we cannot simply increase the number of values to sacrifice efficiency for effect, which contradicts the original intention of GQA. We propose the inner function attention and dynamic attention mask method, which modifies the Value in QCAttn from a linear mapping to the input to a heuristic function that can store more information, and prevents excessive information noise from affecting the calculation of the attention score matrix.

\paragraph{Cross Domain.}
% 状态变换的有效性在于其结构与所需要学习的数据之间的匹配程度. 在人类社会中, 知识广泛分布在不同领域中, 这些领域通过共同的通用知识和交叉领域知识相互连接. 我们设计了交叉领域混合专家, 具有存储通用知识的共享参数和存储领域特定知识的专业参数, 并且专业参数在一定范围内共享, 满足交叉领域知识学习的需求. 并且交叉领域混合专家可以大幅提升专家细粒度, 而不会导致计算速度的快速下降.
The effectiveness of state transformation lies in the match between its structure and the data to be learned. In human society, knowledge is widely distributed in different domains, which are interconnected through common general knowledge and cross domain knowledge. We design cross domain mixture of experts (referred to as CDMoE in the following), which has shared parameters for storing general knowledge and professional parameters for storing domain-specific knowledge, and the professional parameters are shared within a certain range to meet the needs of cross domain knowledge learning. And CDMoE can significantly improve the granularity of experts without causing a rapid decline in computational speed.

\paragraph{Architecture Design.}
% 我们将旋转位置编码矩阵作为状态空间对偶矩阵和因果自注意力矩阵的位置编码方法. 考虑到因果自注意力在长序列计算中较慢, 为了扩展模型深度, 在其之前使用多个状态空间对偶矩阵. 在每个状态空间对偶矩阵或因果自注意力矩阵进行序列变换之后, 使用交叉领域混合专家矩阵进行状态变换. 这些矩阵构成了 Wonderful Matrices 架构, 如图~\ref{fig:wonderful_matrices} 所示.
We use the rotary position embedding matrix as the position encoding method for the state space duality matrix and the quadratic causal self-attention matrix. Considering that the quadratic causal self-attention is slow in long sequence calculations, in order to extend the model depth, multiple state space duality matrices are used before it. After the sequence transformation of each state space duality matrix or quadratic causal self-attention matrix, the cross domain mixture of experts matrix is used for state transformation. These matrices form the Wonderful Matrices architecture, as shown in Figure~\ref{fig:wonderful_matrices}.

% 我们在语言建模任务上进行了实验, 包括具体模块的改进验证, 以及整体架构的验证. 这些实验展示了 Wonderful Matrices 架构在处理复杂语言任务时的高效性和有效性.
We empirically evaluate the Wonderful Matrices architecture on the language modeling task, including the improvement verification of specific modules and the verification of the overall architecture. These experiments demonstrate the efficiency and effectiveness of the Wonderful Matrices architecture in handling complex language tasks.

\section{Related Work}
\label{sec:related_work}

\paragraph{Quadratic Causal Self-Attention.}
% 自注意力是一种计算序列中每个元素与所有其他元素之间关联度的机制, 使得每个元素都能"关注"到其他元素. 最重要的注意力变体是二次自注意力.
Self-Attention is a mechanism that computes the relevance scores between each element in the sequence and all other elements, allowing each element to "attend" to other elements. The most important variant of attention is the quadratic self-attention.

\begin{align*}
Y &= \operatorname*{softmax}({QK^\top}) \cdot V
\end{align*}

% 自注意力的一个显著特点是它可以捕获输入序列中任意位置之间的依赖关系, 而不受距离限制, 并且状态随着序列长度的增加而扩展, 这使得它在捕获长序列中的长距离依赖关系方面具有优势.
A notable feature of quadratic self-attention is that it can capture dependencies between any positions in the input sequence, without being limited by distance, and the state expands with the sequence length, which gives it an advantage in capturing long-range dependencies in long sequences.

% 在因果语言建模中, 通常会添加一个因果掩码, 我将在下文中称之为二次自注意力.
In causal language modeling, a causal mask is usually added to it, which We will refer to as QCAttn (Quadratic Causal Self-Attention) in the following.

\paragraph{State Space Duality.}
% 许多注意力变体都是基于注意力分数的核心提出的.
Many variants of attention have been proposed, all of which are based on the core of attention scores.

% 线性注意力~\citep{katharopoulos2020transformers}通过将 softmax 折叠到核特征映射中, 并使用矩阵乘法的核属性将 $(QK^\top) \cdot V = Q \cdot (K^\top V)$ 重写. 在因果(自回归)注意力的情况下, 他们展示了当因果掩码合并到左侧时 $(L \circ QK^\top) \cdot V$, 其中 $L$ 是一个下三角矩阵, 右侧可以展开为一个递归形式.
linear attention~\citep{katharopoulos2020transformers} discards softmax by folding it into the kernel feature map and rewrites $(QK^\top) \cdot V = Q \cdot (K^\top V)$ using the kernel property of matrix multiplication. In the case of causal (autoregressive) attention, they show that when the causal mask is merged to the left as $(L \circ QK^\top) \cdot V$, where $L$ is a lower triangular matrix, the right side can be expanded into a recursive form.

% 在 Transformers are SSMs~\citep{mamba2} 中, 使用 SSD (State Space Dual) 来证明, 简单地计算标量结构化 SSM --- 通过实现半可分矩阵 $M = L \circ CB^\top = L \circ QK^\top$ 并执行二次矩阵-向量乘法 --- 等价于二次掩码核注意力.
In Transformers are SSMs~\citep{mamba2}, the SSD (State Space Dual) is used to prove that simply computing the scalar structured SSM --- by materializing the semi-separable matrix $M = L \circ CB^\top = L \circ QK^\top$ and performing quadratic matrix-vector multiplication --- is equivalent to quadratic masked kernel attention.

\begin{align*}
  \label{eq:attention_ssm}
  (L \circ QK^\top) \cdot V = (L \circ CB^\top) \cdot X
\end{align*}

\paragraph{Positional Encoding.}
% 位置信息在语言建模中很重要, 主要有三种形式的相对位置编码: 卷积、递归和内积.
Position information is important in language modeling, and there are mainly three forms of relative positional encoding: convolution, recursive, and inner product.

% Mamba~\citep{gu2023mamba} 中的位置信息来源是因果卷积和跳过输入和输出之间的连接的矩阵 D.
The source of positional information in Mamba~\citep{gu2023mamba} is causal convolution and matrix D that skips the connection between input and output.

% 在 Mamba2~\citep{mamba2} 中, 元素 $a_t$ 充当一个"门"或"选择器", 其累积乘积 $a_(j:i)$ 控制允许位置 $i$ 和位置 $j$ 之间的交互的量, 这可以看作是一种相对位置嵌入的形式.
In Mamba2~\citep{mamba2}, element $a_t$ acts as a "gate" or "selector", and its cumulative product $a_(j:i)$ controls the amount of interaction allowed between position $i$ and position $j$, which can be seen as a form of relative positional embedding.

% RoPE~\citep{su2021roformer} 在自注意力中添加绝对位置信息, 并通过计算 $QK^\top$ 的内积得到相对位置编码矩阵 $[B, L, L]$.
RoPE~\citep{su2021roformer} adds absolute positional information to $Q$ and $K$ in self-attention, and obtains the relative positional encoding matrix $[B, L, L]$ by calculating the inner product of $QK^\top$ $[B, L, D] \times [B, L, D]^\top$.

\paragraph{Mixture of Experts.}
% 稀疏激活的混合专家架构旨在在有限的计算资源下更少的训练步骤中训练一个更大的模型, 这通常比在更多步骤中训练一个更小的模型表现更好.
The sparse activation mixture of experts architecture aims to train a larger model in fewer training steps with limited computational resources, which often performs better than training a smaller model in more steps.

% 在路由专家策略中, 为了确保专家学习到不重复的通用知识, shared expert isolation通过隔离 k 个专家进行知识共享, 将隔离专家的整个序列状态添加到路由专家每个 token 的状态中, 以确保专家学习到不重复的通用知识.
In the routing expert strategy, to ensure that the experts learn non-redundant general knowledge, shared expert isolation~\citep{dai2024deepseekmoe}(referred to as SEI in the following) shares knowledge by isolating k experts, adding the entire sequence state of the isolated experts to the state of each token of the routing expert to ensure that the experts learn non-redundant general knowledge.

\paragraph{Mixture of A Million Experts.}
% 稀疏激活的混合专家的粒度越细, 性能越好. Mixture of A Million Experts~\citep{he2024moame} 提出了 PEER (parameter efficient expert retrieval) 来在大量专家下保持计算效率.
The finer the granularity of the sparse activation mixture of experts, the better the performance. Mixture of A Million Experts~\citep{he2024moame} proposes PEER (parameter efficient expert retrieval) to maintain computational efficiency under a large number of experts.

\paragraph{Expressive Hidden States.}
% Learning to (Learn at Test Time)~\citep{sun2024ttt} 提出将隐藏状态设置为机器学习模型本身, 以增加其表达能力. 隐藏状态是一个可以训练执行任何任务的神经网络, 模型可以训练学习隐藏状态.
Learning to (Learn at Test Time)~\citep{sun2024ttt} proposes to make the hidden state the machine learning model itself to increase its expressive power. The hidden state is a neural network that can be trained to perform any task, and the model can be trained to learn the hidden state.

\section{Methods}
\label{sec:methods}

% Wonderful Matrices 是一个基础架构, 旨在构建高效和有效的模型.
Wonderful Matrices is a foundation architecture designed to build efficient and effective models.

\paragraph{Rotary Position Embedding for Hybrid Algorithms.}
% 当混合SSD与QCAttn算法时, 我们证明了旋转位置编码矩阵在混合算法中的可用性. 该方法将在第~\ref{sec:rope_for_hybrid_algorithms} 节中描述.
When mixing SSD with QCAttn algorithms, we have proven the availability of the rotary position embedding matrix in the hybrid algorithm. The method is described in Section \ref{sec:methods:rope_for_hybrid_algorithms}.

\paragraph{Inner Function Attention with Dynamic Attention Mask.}
% 我们提出了内函数注意力来存储更多共享的值状态, 并配合动态注意力掩码来保证注意力分数的准确性. 该方法将在第~\ref{sec:methods:ifa} 节中描述.
We propose inner function attention to store more shared value states, and cooperate with dynamic attention mask to ensure the accuracy of attention scores. The method is described in Section \ref{sec:methods:ifa}.

\paragraph{Cross Domain Mixture of Experts.}
% 我们提出了交叉领域混合专家. 共享参数与私有参数的比例可以任意调整, 并且与经典路由混合专家相比, 它扩展大量专家数量, 而不会导致专家检索速度下降. 该方法将在第~\ref{sec:methods:cdmoe} 节中描述.
We propose CDMoE. The ratio of shared parameters to private parameters can be adjusted arbitrarily, and compared with the classic routing mixture of experts, it expands the number of experts significantly without causing a decrease in expert retrieval speed. The method is described in Section \ref{sec:methods:cdmoe}.

\paragraph{Architecture Design.}
% 我们组合这些方法设计了 Wonderful Matrices 架构. 该架构将在第~\ref{sec:methods:architecture} 节中描述.
We combine these methods to design the Wonderful Matrices architecture. The architecture is described in Section \ref{sec:methods:architecture}.

\subsection{Rotary Position Embedding for Hybrid Algorithms}
\label{sec:methods:rope_for_hybrid_algorithms}

% 例如, 在自注意力 $QK^\top$ 中, 计算两个向量 $Q_i \cdot K_j$ 的点积, 得到的结果是一个标量, 代表位置 $i$ 和位置 $j$ 之间的相关性.
For example, in the self-attention $QK^\top$, the dot product of two vectors $Q_i \cdot K_j$ is calculated, and the result is a scalar, which represents the correlation between position $i$ and position $j$.

% 旋转位置编码的基本思想是将位置信息编码为一个复数旋转矩阵, 其角度由位置索引决定. 当 $QK$ 或 $CB$ 与 RoPE 结合时, 如果一个元素位置靠前, 其旋转将影响与之相乘的 $K$ 或 $B$ 向量的方向, 从而影响内积的结果.
\paragraph{Dot Product Rotary Position}
The basic idea of rotary position embedding is to encode the position information as a complex rotary matrix, whose angle is determined by the position index. When $QK$ or $CB$ is applied with RoPE, if an element position is close to the front, its rotary will affect the direction of the $K$ or $B$ vector multiplied by it, thereby affecting the result of the inner product.

% 定义 RoPE 为 $f_{\{Q, K\}}(x_{i}, i)$ 和 $f_{\{C, B\}}(x_{i}, i)$, 其中 $x_{i}$ 是输入向量, $i$ 是位置索引, 则:
Define RoPE as $f_{\{Q, K\}}(x_{i}, i)$ and $f_{\{C, B\}}(x_{i}, i)$, where $x_{i}$ is the input vector, and $i$ is the position index, then:

\begin{align}
    f_{\{Q, K\}}(x_{i}, i) &= \mathbb{R}_{\Theta, i}^{d} W_{\{Q, K\}} x_{i} \\
    f_{\{C, B\}}(x_{i}, i) &= \mathbb{R}_{\Theta, i}^{d} W_{\{C, B\}} x_{i} \label{eq:rope_cb}
\end{align}

% 其中 $\Theta$ 和 $\R_{\Theta, i}^{d}$ 定义如下:
where $\Theta$ and $\mathbb{R}_{\Theta, i}^{d}$ are defined as follows:

\begin{align*}
    \Theta = \{\theta_i = n^{-2(i - 1) / d}, i \in [1, 2, \dots, d / 2]\} \quad
    \mathbb{R}_{\Theta, i}^{d} = 
    \begin{bmatrix}
        \cos i \theta_0 & -\sin i \theta_0 & 0 & 0 & \dots & 0 & 0 \\        
        \sin i \theta_0 & \cos i \theta_0 & 0 & 0 & \dots & 0 & 0 \\        
        0 & 0 & \cos i \theta_1 & -\sin i \theta_1 & \dots & 0 & 0 \\        
        0 & 0 & \sin i \theta_1 & \cos i \theta_1 & \dots & 0 & 0 \\        
        \vdots & \vdots & \vdots & \vdots & \ddots & \vdots & \vdots \\        
        0 & 0 & 0 & 0 & \dots & \cos i \theta_{d/2} & -\sin i \theta_{d/2-1} \\        
        0 & 0 & 0 & 0 & \dots & \sin i \theta_{d/2} & \cos i \theta_{d/2-1} \\    
    \end{bmatrix}
\end{align*}

% 在附录~\ref{sec:rope_for_ssd} 中我们证明了 RoPE 在 SSD 算法中的可用性. 在图~\ref{fig:rope_and_ifa} 中展示了旋转位置编码的算法矩阵. 在附录~\ref{sec:implementation_code:rope} 中提供了 RoPE 的实现代码示例, 以及 RoPE 在 Attn 和 SSD 中的应用.
In Appendix~\ref{sec:rope_for_ssd}, we prove the availability of RoPE in the SSD algorithm. The algorithm matrix of the rotary position embedding is shown in Figure~\ref{fig:rope_and_ifa}. In Appendix~\ref{sec:implementation_code:rope}, an implementation code example of RoPE and its application in Attn and SSD are provided.

\subsection{Inner Function Attention with Dynamic Mask}
\label{sec:methods:ifa}

\begin{figure}[h]
    \centering
    \begin{subfigure}{0.49\linewidth}
        \centering
        \includegraphics[height=0.38\textheight]{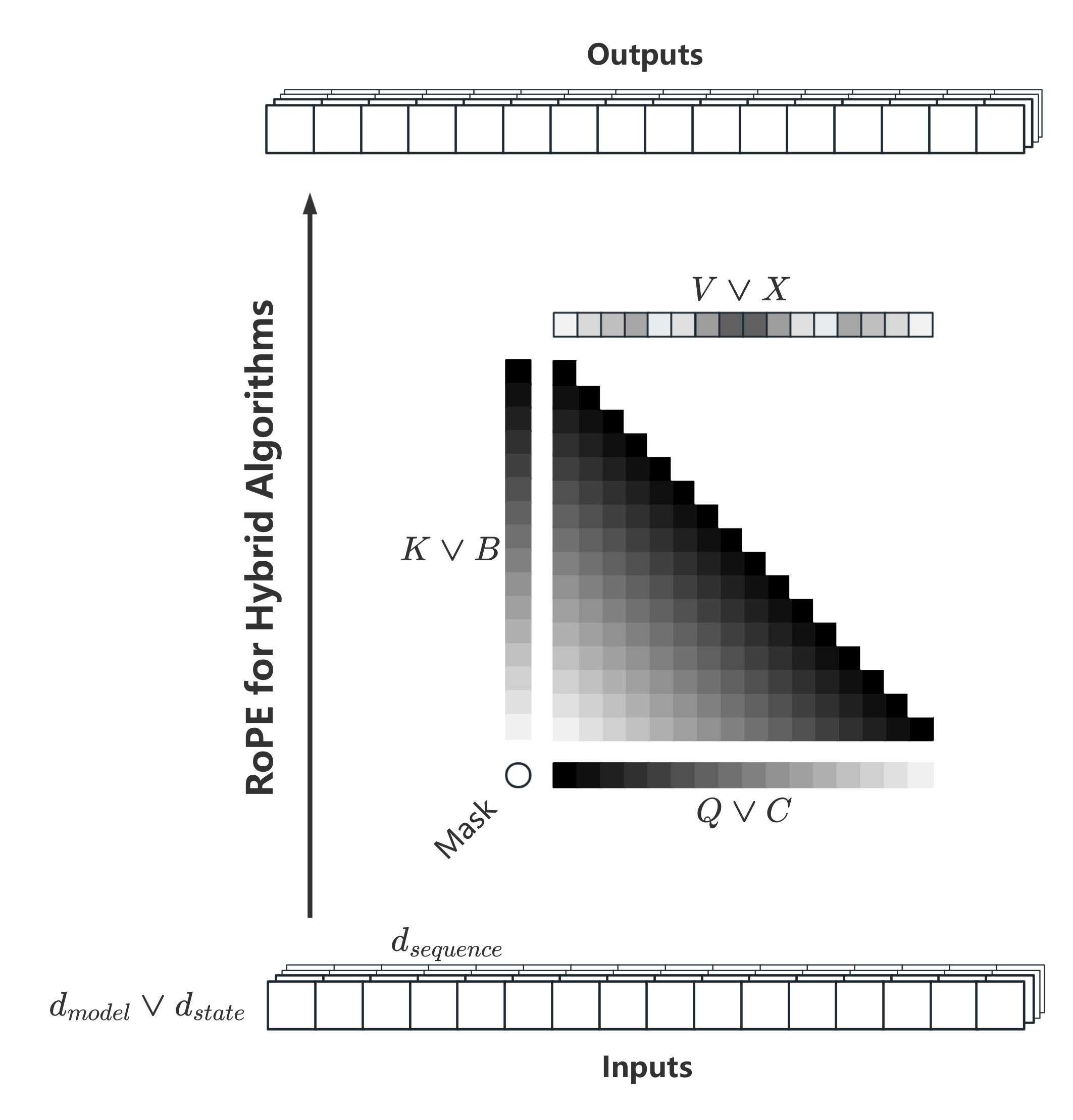}
    \end{subfigure}
    \begin{subfigure}{0.49\linewidth}
        \centering
        \includegraphics[height=0.38\textheight]{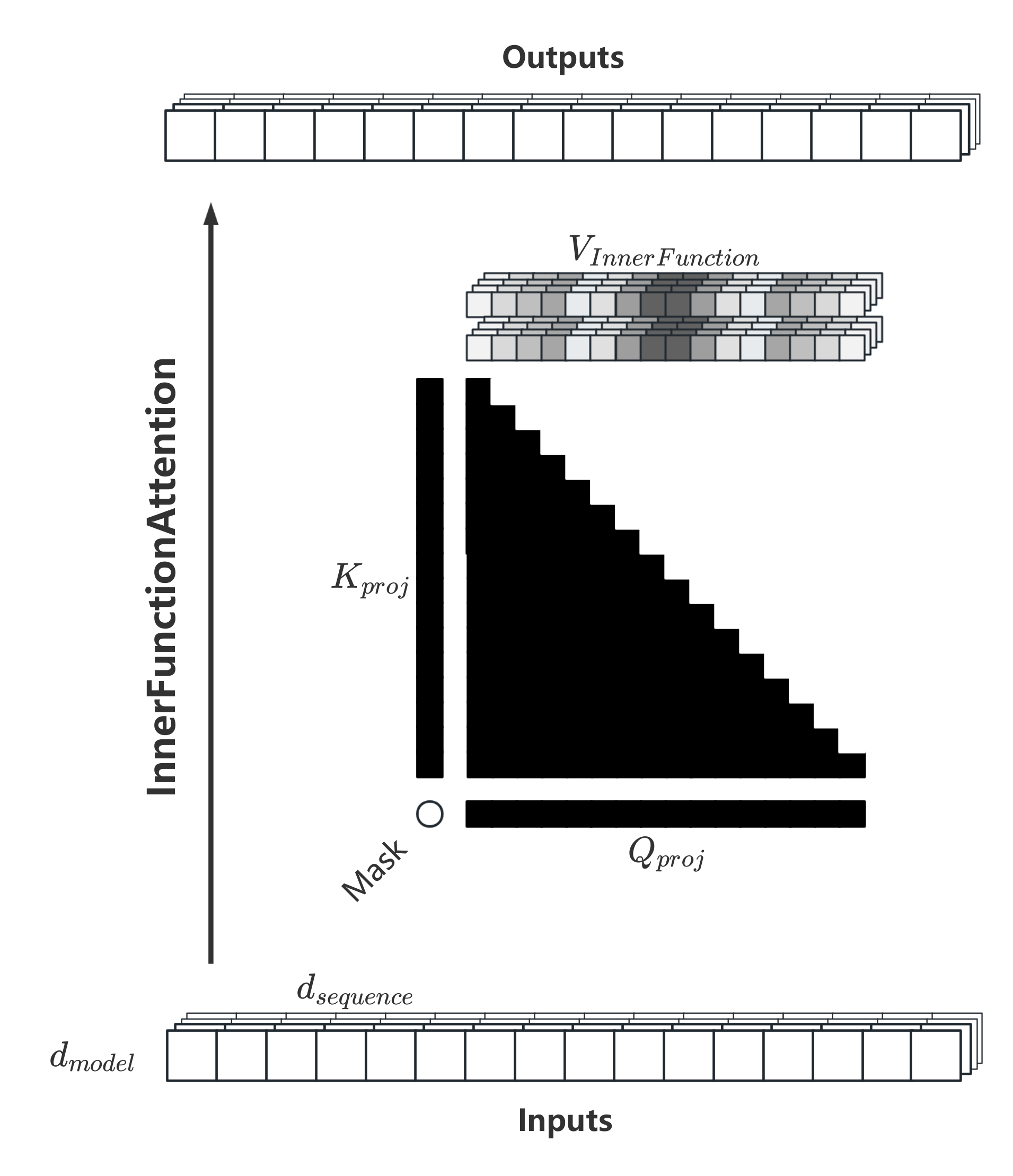}
    \end{subfigure}
    \caption{
        (\textbf{Left})
        \textbf{RoPE for Hybrid Algorithms}.
        % 展示了旋转位置编码在旋转位置编码矩阵中的算法矩阵, 以内积的形式展示了旋转位置编码的算法矩阵.
        % 颜色的深浅代表位置编码的位置, 颜色越深, 位置编码越高.
        % 输入张量首先与 $QK$ 或 $CB$ 矩阵相乘, 然后附加正弦余弦位置信息, 最后通过内积运算得到标量的相对位置矩阵.
        Shows the algorithm matrix of the rotary position embedding in the form of inner product.
        The depth of color represents the position of the position encoding, with higher color depth and lower color depth.
        The input tensor is first multiplied by the $QK$ or $CB$ matrix, then the sine and cosine position information is attached, and finally the relative position matrix of the scalar is obtained through the inner product operation.
        (\textbf{Right})
        \textbf{Inner Function Attention}.
        % 展示了内函数注意力中的结构和算法.
        % 输入张量首先与 $QK$ 相乘得到查询和键矩阵, 然后计算标量注意力分数, 最后将注意力分数附加到内函数计算的值状态上并输出.
        Shows the structure and algorithm of inner function attention.
        The input tensor is first multiplied by $QK$ to obtain the query and key matrix, then the scalar attention score is calculated, and finally the attention score is attached to the value state calculated by the inner function and output.
    }
    \label{fig:rope_and_ifa}
\end{figure}

% 二次自注意力算法计算与整个输入序列 $x$ 相关的 $Q$ 和 $K$ 以形成注意力矩阵, 隐藏状态 (通常称为 $KV$ 缓存) 是一个线性增长的列表, 显式存储所有历史上下文信息而不进行任何压缩, 计算这种线性增长状态的时间复杂度是二次增加.
The QCAttn algorithm calculates the $Q$ and $K$ related to the entire input sequence $x$ to form the attention matrix, and the hidden state (usually referred to as the $KV$ cache) is a linearly growing list with $t$ (token), explicitly storing all historical context information without any compression, and the time complexity of calculating this linearly growing state is quadratically increasing. 

% 为了增强二次自注意力隐藏状态的表达能力, 我们的想法是将部分隐藏状态计算从简单的 $y = xW$ 转变为优秀的启发式 $y = f(x)$. 考虑到 $Q$ 和 $K$ 需要执行内积运算, 更复杂的操作可能会导致训练过程中的大幅波动, 因此启发式应用于 $V$. 当然, 从更直观的角度来看, 查询和键确实只需要简单的线性变换, 它们的点积注意力矩阵, 是确定从值中提取哪些信息的. 因此,我们可以将启发式~\ref{eq:inner_function_V}应用于值以提高隐藏状态的表达能力.
\paragraph{Inner Function.}
To enhance the expressive power of the hidden state of QCAttn, our idea is to transform part of the hidden state calculation from a simple $y = xW$ to an excellent heuristic $y = f(x)$. Considering that $Q$ and $K$ need to perform inner product operations, more complex operations may cause large fluctuations during training, so the heuristic is applied to $V$. Of course, from a more intuitive perspective, queries and keys do only need simple linear transformations, and their dot product attention matrix determines which information to extract from the values. Therefore, we can apply the heuristic~\ref{eq:inner_function_V} to the value to improve the expressive power of the hidden state.

\begin{align}
    V_{innerfunc} = f(x) := (xW_{vQ}\theta_{vK}^T)W_{V}^{T} \times x
    \label{eq:inner_function_V}
\end{align}

% 我们首先初始化参数 $W_{Q}$, $\theta_{K}$, $W_{V}$. 其中 $W_{Q}$ 和 $\theta{K}$ 还非常易于扩展为 $\mathbb{R}^{d_{model} \times n_{heads} \times d_{ret}}$ 和 $\mathbb{R}^{n_{v} \times n_{heads} \times d_{ret}}$ 这种多头形式, 但是为了简单起见我们只展示基础计算框架. 其中 $d_{model}$ 代表模型隐藏维度, $d_{ret}$ 代表检索状态维度, $n_{v}$ 代表值的数量.
First, We initialize the parameters $W_{vQ}$, $\theta_{vK}$, $W_{V}$. Where $W_{vQ}$ and $\theta{vK}$ are also very easy to expand into the multi-head, but for simplicity, we only show the basic computational framework. Where $d_{model}$ represents the model hidden dimension, $d_{ret}$ represents the retrieval dimension, $n_{v}$ represents the number of values.

\begin{align*}
    W_{vQ} &\in \mathbb{R}^{d_{model} \times d_{ret}} \quad \theta_{vK} \in \mathbb{R}^{n_{v} \times d_{ret}} \quad W_{V} \in \mathbb{R}^{n_{v} \times d_{model}} \\
\end{align*}

% 第二步使输入状态 $x \in \mathbb{R}^{batch \times seq \times d_{model}}$ 与 $W_{Q}$ 进行矩阵乘法, 得到输入的低秩投影, 然后与 $\theta_{K}^T$ 进行点积计算, 得到相似度标量分数 $g \in \mathbb{R}^{batch \times seq \times n_{v}}$.
The second step is to perform matrix multiplication of the input state $x \in \mathbb{R}^{batch \times seq \times d_{model}}$ with $W_{vQ}$ to obtain the low-rank projection of the input, and then perform dot product calculation with $\theta_{vK}^T$ to obtain the similarity scalar score $g$.

\begin{align*}
    % 计算相似度
    g &= x \cdot W_{vQ} \cdot \theta_{vK}^T \in \mathbb{R}^{batch \times seq \times n_{v}}
\end{align*}

% 第三步在 $g$ 的 $n_{v}$ 维度取 topk 个 values 对应的索引, 得到 $i \in \mathbb{R}^{batch \times seq \times k}$. 值得注意的是, 输出多样性随着 $k$ 的增加而增加.
The third step is to take the index of the topk values corresponding to the $n_{v}$ dimension of $g$, obtaining $i$. It is worth noting that the output diversity increases with the increase of $k$.

\begin{align*}
    % 取在最后n_v维度取topk个value对应的索引
    i &= \operatorname*{topk}(g, k, dim=-1).indices \in \mathbb{R}^{batch \times seq \times k}
\end{align*}

% 第四步取出索引位置 $i$ 对应的 $W_{V}^T$ 的 values 状态, 得到 $v \in \mathbb{R}^{batch \times seq \times k \times d_{model}}$.
The fourth step is to take out the values state of $W_{V}^T$ corresponding to the index position $i$, obtaining $v$.

\begin{align*}
    % 取出对应的values
    v &= i \cdot W_{V}^T \in \mathbb{R}^{batch \times seq \times k \times d_{model}}
\end{align*}

% 最后将 $v$ 在 $k$ 维度叠加并与输入状态 $x$ 相关联, 得到 $V \in \mathbb{R}^{batch \times seq \times d_{model}}$.
Finally, the values $v$ are superimposed in the $k$ dimension and associated with the input state $x$, obtaining $V$.

\begin{align*}
    % 使values状态叠加并与输入状态相关联
    % V &= x \times \sum^{k}_{i=1} v \in \mathbb{R}^{batch \times seq \times d_{model}}
    V &= \sum^{k}_{i=1} x \times v_i \in \mathbb{R}^{batch \times seq \times d_{model}}
\end{align*}

% 我们为每个token都分配与其亲和度最高的topk的values, 并且所有values的权重都是共享的行, 不会出现稀疏结构导致的训练不足问题. 值得注意的是, 这种内函数算法
We assign the topk values with the highest affinity to each token, and all the weights of the values are shared rows, avoiding the problem of insufficient training caused by the sparse structure.

% 为了保留QCAttn对所有历史上下文信息的显式存储, 我们仍然使用 $QK^T$ 计算的注意力矩阵为状态 $V_{innerfunc}$ 分配权重分数. 然而注意力矩阵分数的精度主要取决于 $n_{head}$ 的大小与 $softmax$ 函数的计算精度. 在十分长的上下文中, 较小的 $n_{head}$ 会导致模型迷失在上下文中, 而较大的 $n_{head}$ 会导致计算量问题.
To retain the explicit storage of all historical context information by QCAttn, we still use the attention matrix calculated by $QK^T$ to assign weight scores to the state $V_{innerfunc}$. However, the accuracy of the attention matrix score depends mainly on the dimension of $n_{head}$ and the calculation accuracy of the $softmax$ function. In very long contexts, a smaller $n_{head}$ will cause the model lost in the middle~\citep{liu2023lostmiddlelanguagemodels}, while a larger $n_{head}$ will cause computational problems. 

\begin{align*}
    W_{Q} &\in \mathbb{R}^{d_{model} \times n_{heads} \times d_{head}} \quad W_{K} \in \mathbb{R}^{d_{model} \times n_{heads} \times d_{head}} \\
    Q &= xW_Q \in \mathbb{R}^{batch \times n_{heads} \times seq \times d_{head}} \\
    K &= xW_K \in \mathbb{R}^{batch \times n_{heads} \times seq \times d_{head}} \\
    A &= softmax(QK^T) \in \mathbb{R}^{batch \times n_{heads} \times seq \times seq} \\
\end{align*}

% 一般来说, 我们训练transformers模型都会使用tokenizer将序列填充为一个固定的长度, 以便于批量训练. 填充部分的注意力掩码我们使用 $0$ 来表示, 以便于在计算注意力分数时忽略这部分的信息. 而原始序列部分的注意力掩码通常全部表示为 $1$, 表示这部分的信息都是有效的. 这种静态注意力掩码处理方式如附录~\ref{sec:implementation_code:innerfuncattn}中的实现示例\ref{lst:static_mask}所示.
In general, we train transformers models using a tokenizer to pad the sequence to a fixed length for batch training. We use $0$ to represent the attention mask of the padding part to ignore the information in this part when calculating the attention score. The attention mask of the original sequence part is usually all represented as $1$, indicating that the information in this part is all valid. This static attention mask processing method is shown in the implementation example~\ref{lst:static_mask} in Appendix~\ref{sec:implementation_code:innerfuncattn}.

% SSD 算法使用 $A \in \mathbb{R}^{n_{heads}$ 对上一个状态进行选择性过滤, 来防止无效信息影响到当前状态. 我们模仿这种选择性机制, 在不改动社区发展方向的前提下, 提出了一种简单的即插即用的动态掩码方法. 我们将创建因果掩码的函数移动到注意力层中, 增加一个可以学习的参数\ref{eq:dynamic_mask}, 与因果掩码相乘, 原本填充为 $0$ 的部分不会改变, 又有效部分会随着动态掩码参数的不断学习而掩盖其中的噪声, 如附录~\ref{sec:implementation_code:innerfuncattn}中的实现示例\ref{lst:dynamic_mask}所示. 这种动态掩码的方式同时也随着注意力层的堆叠而堆叠, 从而满足序列变换的状态选择性.
\paragraph{Dynamic Attention Mask.}
The SSD algorithm uses $A \in \mathbb{R}^{n_{heads}}$ to selectively filter the previous state to prevent invalid information from affecting the current state. We imitate this selective mechanism and propose a simple plug-and-play dynamic mask method without changing the direction of community development. We move the function of creating the causal mask to the attention layer, add a learnable parameter~\ref{eq:dynamic_mask}, multiply it with the causal mask, the part originally filled with $0$ will not change, and the effective part will be covered with noise as the dynamic mask parameter continues to learn, as shown in the implementation example~\ref{lst:dynamic_mask} in Appendix~\ref{sec:implementation_code:innerfuncattn}. This dynamic mask method is also stacked with the attention layer as the attention layer is stacked, thereby satisfying the selectivity of the sequence transformation state.

\begin{align}
    l &:= \operatorname{round}(\operatorname{ones}([n_{heads}, max\_position\_len])) 
    \label{eq:dynamic_mask}
\end{align}

% 最后我们将处理后的注意力掩码应用在注意力分数矩阵上, 以保证从值中提取的信息是有效的. 
Finally, we apply the processed attention mask to the attention score matrix to ensure that the information extracted from the values is valid.

\begin{align*}
    L &= \operatorname{dynamic\_mask}(l) \\
    M &= L \circ A \\
    y &= M \cdot V_{innerfunc}
\end{align*}

% 这种内函数注意力配合动态注意力掩码的方式, 不仅可以有效扩展序列变换的隐藏状态表达能力, 还可以减少标量关注分数的误差. 在图~\ref{fig:rope_and_ifa} 中展示了内函数注意力的算法矩阵. 在附录~\ref{sec:implementation_code:ssd}和\ref{sec:implementation_code:ssdattn}中提供了 SSD 的实现代码示例. 在附录~\ref{sec:implementation_code:innerfuncattn}中提供了内函数注意力的实现代码示例.
This inner function attention with dynamic attention mask can not only effectively expand the hidden state expressive power of sequence transformation but also reduce the error of scalar attention scores. The algorithm matrix of inner function attention is shown in Figure~\ref{fig:rope_and_ifa}. An implementation code example of SSD is provided in Appendices~\ref{sec:implementation_code:ssd}. An implementation code example of inner function attention is provided in Appendix~\ref{sec:implementation_code:innerfuncattn}.

\subsection{Cross Domain Mixture of Experts}
\label{sec:methods:cdmoe}

\begin{figure}[ht]
    \centering
    \includegraphics[width=\linewidth]{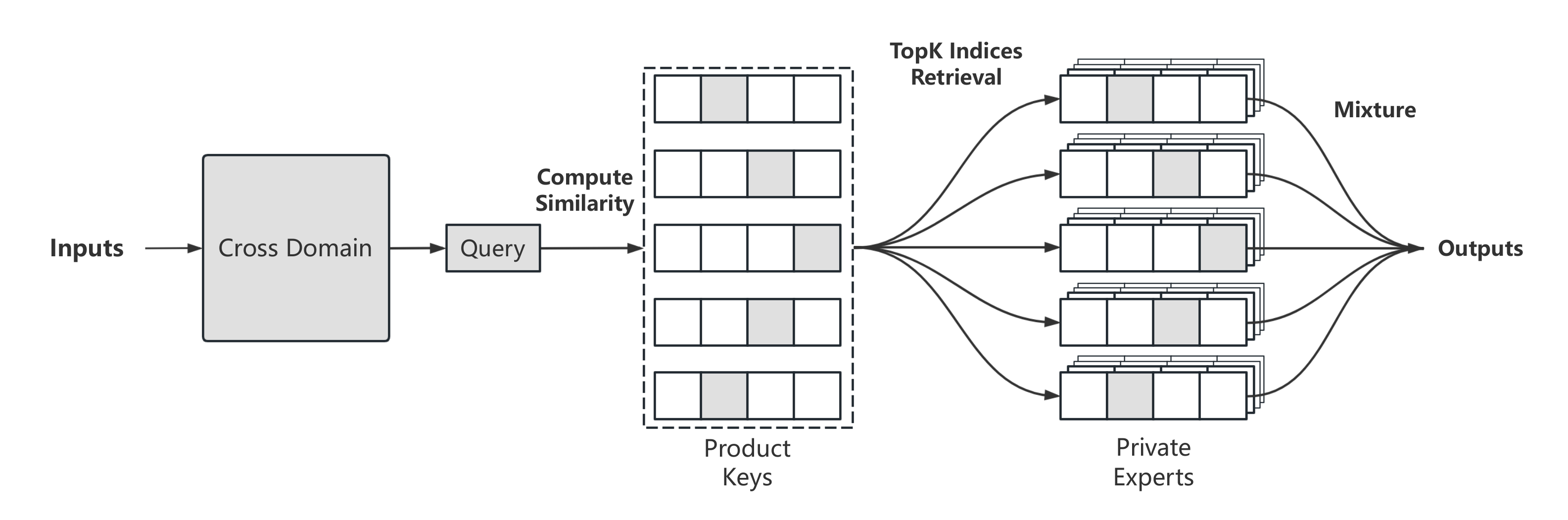}
    \caption{
        \textbf{CDMoE}.
        % 展示了交叉领域百万混合专家矩阵的内部结构和计算过程. 输入张量首先通过交叉领域的共享参数, 然后通过线性层并重塑为查询, 然后与键计算点积以获取与私有专家的亲和度, 最后通过具有最高亲和度的前 K 个私有专家混合携带共享知识的张量.
        Shows the internal structure and calculation process of the cross domain million mixture of experts matrix.
        Input tensors first pass through the shared parameters of the cross domain, then pass through a linear layer and reshape into queries, then calculate the dot product with the keys to obtain the affinity with the private experts, and finally mix the tensors carrying shared knowledge through the top K private experts with the highest affinity.
    }
    \label{fig:cdmoe}
\end{figure}

% 在传统的混合专家策略中, 分配给不同专家的令牌需要具有共同的知识或信息, 因此当获取各自的参数时, 多个专家将具有存储共同信息的冗余参数, 从而导致专家参数冗余. 专家参数冗余的比例随着专家粒度的增加和专家激活数量的减少而增加. 在 Mixture of A Million Experts~\citep{he2024moame} 中, 由于专家精细度高, 即使专家权重行全部共享, 仍然因为专家数量列激活数量低导致专家参数冗余的比例达到了惊人的高度, 可以从预训练开始的损失曲线中看出. 所以我们在此基础上提出了两种形式方程的交叉领域混合专家~\ref{eq:cdmoe}, 其中 $e$ 代表专家, $N$ 代表专家数量.
In the conventional mixture of experts strategy, the tokens assigned to different experts need to have common knowledge or information, so multiple experts will have redundant parameters for storing common information when obtaining their respective parameters, which leads to expert parameter redundancy. And the proportion of expert parameter redundancy increases with the increase in expert granularity and the decrease in the number of expert activations. In Mixture of A Million Experts~\citep{he2024moame}, due to the high granularity of the experts, even if all the expert weight rows are shared, the proportion of expert parameter redundancy reaches an astonishingly high level due to the low number of expert column activations, as can be seen from the loss curve starting from pre-training. So we propose a cross domain mixture of experts for two forms of equations~\ref{eq:cdmoe}, where $e$ represents the expert, and $N$ represents the number of experts.

\begin{align}
    e(x) = \sum_{i=1}^{N} e_i(x) + \phi(x) \quad e(x) = \sum_{i=1}^{N} e_i(\phi(x))
    \label{eq:cdmoe}
\end{align}

% 如果分配给不同专家的令牌已经通过存储共同知识的参数, 则可以减少参数冗余. 这种存储共同知识的参数可以称为交叉领域. 交叉领域的输出作为私有混合专家的输入, 而混合专家都会有一种亲和度计算策略来决定使用哪个专家, 或者是路由遍历策略, 或者点积分数索引策略. 这种亲和度计算就相当于一个门控机制, 用于决定哪个token应该由哪个专家处理, 所以我们没有使用门控MLP作为交叉领域. 其中 $s$ 代表共享参数, $\sigma$ 代表激活函数, $W, V$ 代表MLP的两个权重矩阵, $d_{model}$ 代表模型隐藏维度, $d_{ff}$ 代表前馈网络的表示维度.
\paragraph{Cross Domain.}
If the tokens assigned to different experts have already passed through the parameters for storing common knowledge, the parameter redundancy can be reduced. This parameter for storing common knowledge can be called cross domain.The output of the cross domain is used as the input of the private mixture of experts, and the mixture of experts will have an affinity calculation strategy to determine which expert to use, either a routing traversal strategy or a dot product score index strategy. This affinity calculation is equivalent to a gating mechanism used to determine which token should be processed by which expert, so we do not use a gated MLP as the cross domain. where $s$ represents shared parameters, $\sigma$ represents the activation function, $W, V$ represents the two weight matrices of the MLP, $d_{model}$ represents the model hidden dimension, and $d_{ff}$ represents the representation dimension of the feedforward network.

\begin{align*}
    \phi(x) &= \sigma(xW^s)V^s \quad W^s \in \mathbb{R}^{d_{model} \times d_{ff}} \quad V^s \in \mathbb{R}^{d_{ff} \times d_{model}}
\end{align*}

% 现在我们列举一种高效检索的混合专家 $e(x)$ 算法来完成形式方程二的计算, 因为在形式二的基础上只需要调整一下计算参数即可完成形式方程一. 我们首先初始化查询投影矩阵权重 $W_{eQ}$, 可以学习的键参数 $\theta_{eK}$, 两个权重矩阵 $W_{edown}$ 和 $W_{eup}$. 其中 $p$ 代表私有参数, $d_{ret}$ 代表专家检索状态维度, $n_{expert}$ 代表专家数量, $n_{h}$ 代表头数, $k$ 代表 topk 的数量.
\paragraph{Efficient Retrieval Experts.}
Now we list an efficient retrieval expert $e(x)$ algorithm to complete the calculation of form equation two, because based on form two, you only need to adjust the calculation parameters to complete form equation one. First, we initialize the query projection matrix weight $W_{eQ}$, learnable key parameters $\theta_{eK}$, and two weight matrices $W_{edown}$ and $W_{eup}$. where $p$ represents the private parameters, $d_{ret}$ represents the expert retrieval state dimension, $n_{expert}$ represents the number of experts, $n_{h}$ represents the number of heads, and $k$ represents the number of topk.

\begin{align*}
    W_{eQ}^{p} \in \mathbb{R}^{d_{model} \times n_{h} \times d_{ret}} \quad \theta_{eK}^{p} \in \mathbb{R}^{n_{h} \times \sqrt{n_{expert}} \times d_{ret}} \quad W_{edown}^{p} \in \mathbb{R}^{n_{expert} \times d_{model}} \quad W_{eup}^{p} \in \mathbb{R}^{n_{expert} \times d_{model}}
\end{align*}

% 第二步首先将交叉领域状态信息$\phi(x)$ 与 $W_{eQ}^{p}$ 进行矩阵乘法运算, 得到共享知识的查询投影, 然后和 $\theta_{eK}^{Tp}$ 进行矩阵乘法得到点积相似度 $g$.
The second step first performs matrix multiplication of the cross domain state information $\phi(x)$ with $W_{eQ}^{p}$ to obtain the query projection of the shared knowledge, and then performs matrix multiplication with $\theta_{eK}^{pT}$ to obtain the dot product similarity $g$.

\begin{align*}
    g &= \phi(x) \cdot W_{eQ}^{p} \cdot \theta_{eK}^{pT} \in \mathbb{R}^{2 \times batch \times seq \times n_{h} \times \sqrt{n_{expert}}}
\end{align*}

% 第三步在 $g$ 的 $\sqrt{n_{expert}}$ 维度为每个 $h$ 取 topk 个专家对应的分数与索引, 经过一些简单的组合步骤, 得到标量分数 $s$ 和专家索引 $i$.
The third step takes the topk experts corresponding to each $h$ in the $\sqrt{n_{expert}}$ dimension of $g$ to obtain the scores and indices, and after some simple combination steps, obtains the scalar scores $s$ and expert indices $i$.

\begin{align*}
    s, i &= \operatorname*{topk}(g, k, dim=-1) \in \mathbb{R}^{batch \times seq \times n_{h} \times k}
\end{align*}

% 第四步将索引位置 $i$ 与 $W_{edown}^{pT}$ 和 $W_{eup}^{pT}$ 进行矩阵乘法运算, 这里与embedding扩展隐藏维度类似, 取出专家维度的权重行, 得到两个索引位置对应的私有状态张量 $d$ 和 $u$.
The fourth step performs matrix multiplication of the index position $i$ with $W_{edown}^{pT}$ and $W_{eup}^{pT}$, similar to embedding to expand the hidden dimension, taking out the weight rows of the expert dimension, obtaining the private state tensors $d$ and $u$ corresponding to the two index positions.

\begin{align*}
    d, u &= i \cdot W_{edown}^{pT}, i \cdot W_{eup}^{pT} \in \mathbb{R}^{batch \times seq \times n_{h} \times k \times d_{model}} \\
\end{align*}

% 第五步我们将交叉领域状态 $\phi(x)$ 与 $d^T$ 进行矩阵乘法计算, 然后与分数 $s$ 相乘并进行非线性激活, 得到专家分数 $s(x)$.
The fifth step performs matrix multiplication of the cross domain state $\phi(x)$ with $d^T$, then multiplies by the score $s$ and non-linearly activates it to obtain the experts score $s(x)$.

\begin{align*}
    s(x) &= \sigma(\phi(x) \cdot d^T \times s) \in \mathbb{R}^{batch \times seq \times n_{h} \times k}
\end{align*}

% 最后, 我们先将 $x$ 与 $u$ 在 $n_{h}, k$ 维度求和, 然后将专家分数 $s(x)$ 与 $u$ 进行矩阵乘法, 得到私有专家的状态, 如果担心上一步的非线性激活会破坏交叉领域状态的话, 则与 $\phi(x)$ 相加, 得到最终的输出 $y$. 
Finally, we first sum $x$ and $u$ in the $n_{h}, k$ dimensions, then perform matrix multiplication of the expert score $s(x)$ with $u$, obtain the state of the private expert, and if you are worried that the non-linear activation in the previous step will destroy the cross domain state, then add it to $\phi(x)$ to obtain the final output $y$.

\begin{align*}
    y &= \sum^{n_{h}}_{i=1} \sum^{k}_{j=1} s(x)_{i, j} \cdot u_{i, j} + \phi(x) \in \mathbb{R}^{batch \times seq \times d_{model}}
\end{align*}

% 这种交叉领域配合高效检索混合专家的方式, 不仅具有一个大型的MLP存储主要的状态变换信息, 还能够动态地将不同小型MLP在 $n_{h}$ 维度组合在一起. 小型MLP通过聚合从共享的权重行中检索到的 $h$ 单例实现共享神经元, 它不仅能够高效的为每个token检索亲和度最高的专家, 并且能够在专家数量增加时保持速度不会像路由策略一样快速下降. 在图~\ref{fig:cdmoe} 中展示了交叉领域混合专家矩阵的内部结构和计算过程. 在附录~\ref{sec:implementation_code:cdmoe}中提供了 CDMoE 的实现代码示例.
This cross domain with efficient retrieval mixture of experts method not only has a large MLP to store the main state transformation information but also can dynamically combine different small MLPs in the $n_{h}$ dimension. The small MLPs share neurons by aggregating the $h$ singletons retrieved from the shared weight rows, which can efficiently retrieve the expert with the highest affinity for each token and maintain speed as the number of experts increases without rapid decline as in the routing strategy. The internal structure and calculation process of the cross domain mixture of experts matrix are shown in Figure~\ref{fig:cdmoe}. An implementation code example of CDMoE is provided in Appendix~\ref{sec:implementation_code:cdmoe}.

\subsection{Architecture Design}
\label{sec:methods:architecture}

\begin{figure}[ht]
  \centering
  \includegraphics[width=.9\linewidth]{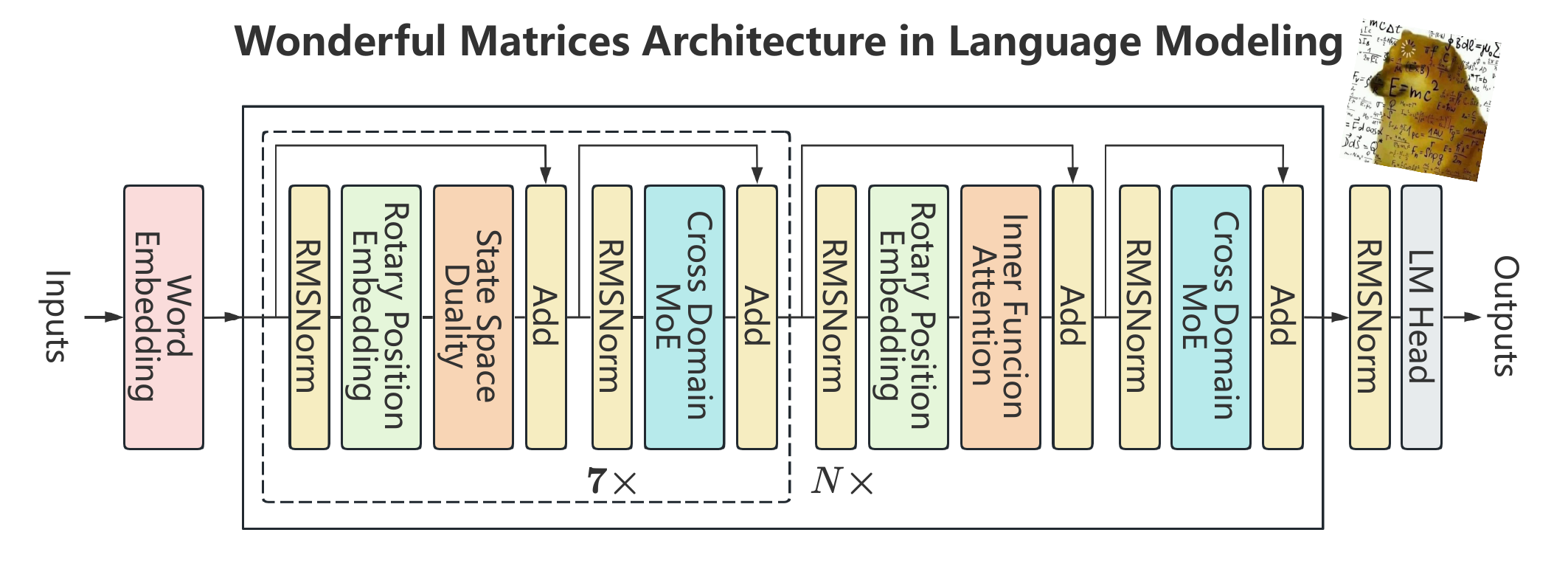}
  \caption{
    \textbf{Wonderful Matrices in Language Modeling: Cheems}.
    % 展示了 Wonderful Matrices 应用在语言建模的架构, 包括词嵌入, RMSNorm, Add(残差连接), RoPE, SSD, InnerFuncAttn, CDMoE, LM Head 模块. 黑色箭头表示模块的计算顺序, 黑色虚线部分表示堆叠该部分 $7$ 次, 黑色实线表示堆叠整个骨干模块部分 $N$ 次. 右上角的狗子是网红柴犬 Cheems, 它算是我们的恶趣味, 让我们在严格的公式推导工作中放松地会心一笑. 为了部分表格的美观, 在后续实验中, 我们将使用 Cheems 作为我们的模型名称.
    Shows the architecture of Wonderful Matrices applied in language modeling, including word embeddings, RMSNorm, Add (residual connection), RoPE, SSD, InnerFuncAttn, CDMoE, and LM Head modules. The black arrows indicate the calculation order of the modules, the black dashed lines indicate stacking this part $7$ times, and the black solid lines indicate stacking the entire backbone module part $N$ times. The dog in the upper right corner is the internet-famous Shiba Inu Cheems, which is our sense of humor, allowing us to relax and smile in strict formula derivation work. For the beauty of the partial table, in subsequent experiments, we will use Cheems as our model name.
  }
  \label{fig:lm_architecture}
\end{figure}

% 我们设计了使用这些矩阵在语言建模任务中的架构: Cheems. 它在开始使用词嵌入将离散词汇转换为连续向量, 并在最后经过final norm之后通过LM Head输出词汇概率分布. 在模型骨干部分, 我们在每一个序列变换模块之前使用RoPE作为位置信息来源, 并在序列变换之后使用一个CDMoE模块作为状态变换, 其中每一个序列变换和状态变换之间都包含输入归一化和残差连接. 在序列变换组合方式中, 我们每堆叠 $7$ 个 SSD 模块, 就堆叠 $1$ 个 InnerFuncAttn 模块(堆叠比例来源于Transformers are SSMs), 以此来保证模型在多查询联系召回任务中的性能. 在图~\ref{fig:lm_architecture} 中展示了 Cheems 的架构.
We designed an architecture using these matrices in the language modeling task: Cheems. In the model backbone, we use RoPE as the source of positional information before each sequence transformation module, and use a CDMoE module as the state transformation after the sequence transformation, with input normalization and residual connections between each sequence transformation and state transformation. In the sequence transformation combination method, we stack $7$ SSD modules for each InnerFuncAttn module stacked (the stacking ratio comes from Transformers are SSMs~\citep{mamba2}), to ensure the model's performance in multi-query contact recall tasks. The architecture of Cheems is shown in Figure~\ref{fig:lm_architecture}.

\section{Empirical Validation}
\label{sec:experiments}

\subsection{Effect of Modules}

\begin{table}[!ht]
    \centering
    \begin{minipage}{.45\linewidth}
        \centering
        \captionsetup{type=table}
        \caption{
        \textbf{$Conv1d + D$ vs. $a_t$ vs. $RoPE$}.
        % 在这里除了上文的简称, 我们还使用 IFA 代表 InnerFuncAttn 模块, DM 代表动态掩码. 单个模块中, QCAttn 无法使用 $Conv1d + D$ 和 $a_t$ . 在设置序列长度为 8192 的情况下, 所有组合模块的困惑表现, $RoPE$ 都优于 $Conv1d$ + D 和 $a_t$.
        In addition to the abbreviations mentioned above, we also use IFA to represent the InnerFuncAttn module and DM to represent the dynamic mask. In a single module, QCAttn cannot use $Conv1d + D$ and $a_t$. With the sequence length set to 8192, the perplexity performance of all combined modules shows that $RoPE$ is better than $Conv1d + D$ and $a_t$.
        }
        \label{tab:comparative_pe}
        \begin{tabular}{@{}ccccccccccc@{}}
        \toprule
        \sc{Modules} & \sc{$Conv1d + D$} & \sc{$a_t$} & \sc{$RoPE$} \\
        & \sc{ppl $\downarrow$} & \sc{ppl $\downarrow$} & \sc{ppl $\downarrow$} \\
        \midrule
        QCAttn & --- & --- & 8.38 \\
        SSD & 8.56 & 8.62 & 8.33 \\
        SSD + QCAttn & 8.48 & 8.56 & 8.18 \\
        SSD + IFA & 8.43 & 8.49 & 8.12 \\
        SSD + IFA + DM & 8.36 & 8.42 & 7.96 \\
        \bottomrule
        \end{tabular}
    \end{minipage}
    \hfill
    \begin{minipage}{.45\linewidth}
        \centering
        \captionsetup{type=table}
        \caption{
            \textbf{MLP vs. CDMoE}.
            % 在这里我们用 S 代表 SSD 模块, A 代表 QCAttn 模块, M 代表 MLP 模块, E 代表 CDMoE 模块. 我们严格构造了相同参数量的不同模型, 在预训练子集上的困惑, 随着 MoE 比例增加而逐渐降低.
            We use S to represent the SSD module, A to represent the QCAttn module, M to represent the MLP module, and E to represent the CDMoE module. We strictly construct different models with the same number of parameters, and the perplexity on the pre-training subset gradually decreases as the MoE ratio increases.
        }
        \label{tab:comparative_ffn}
        \begin{tabular}{@{}ccccccccccccccccc@{}}
        \toprule
        \sc{Modules} & \sc{MoE Ratio} & \sc{ppl $\downarrow$} \\
        \midrule
        \makebox[1.25em]{\textbf{SM}} \makebox[1.25em]{\textbf{SM}} \makebox[1.25em]{\textbf{SM}} \makebox[1.25em]{\textbf{SM}} \makebox[1.25em]{\textbf{SM}} \makebox[1.25em]{\textbf{SM}} \makebox[1.25em]{\textbf{SM}} \makebox[1.25em]{\textbf{AM}} & 0\% & 8.18 \\
        \makebox[1.25em]{\textbf{SE}} \makebox[1.25em]{\textbf{SM}} \makebox[1.25em]{\textbf{SM}} \makebox[1.25em]{\textbf{SM}} \makebox[1.25em]{\textbf{SM}} \makebox[1.25em]{\textbf{SM}} \makebox[1.25em]{\textbf{SM}} \makebox[1.25em]{\textbf{AM}} & 6.25\% & 8.06 \\
        \makebox[1.25em]{\textbf{SM}} \makebox[1.25em]{\textbf{SM}} \makebox[1.25em]{\textbf{SM}} \makebox[1.25em]{\textbf{SM}} \makebox[1.25em]{\textbf{SM}} \makebox[1.25em]{\textbf{SM}} \makebox[1.25em]{\textbf{SM}} \makebox[1.25em]{\textbf{AE}} & 6.25\% & 8.12 \\
        \makebox[1.25em]{\textbf{SE}} \makebox[1.25em]{\textbf{SM}} \makebox[1.25em]{\textbf{SM}} \makebox[1.25em]{\textbf{SM}} \makebox[1.25em]{\textbf{SM}} \makebox[1.25em]{\textbf{SM}} \makebox[1.25em]{\textbf{SM}} \makebox[1.25em]{\textbf{AE}} & 12.5\% & 7.96 \\
        \makebox[1.25em]{\textbf{SM}} \makebox[1.25em]{\textbf{SE}} \makebox[1.25em]{\textbf{SE}} \makebox[1.25em]{\textbf{SE}} \makebox[1.25em]{\textbf{SE}} \makebox[1.25em]{\textbf{SE}} \makebox[1.25em]{\textbf{SE}} \makebox[1.25em]{\textbf{AM}} & 37.5\% & 7.52 \\
        \makebox[1.25em]{\textbf{SE}} \makebox[1.25em]{\textbf{SE}} \makebox[1.25em]{\textbf{SE}} \makebox[1.25em]{\textbf{SE}} \makebox[1.25em]{\textbf{SE}} \makebox[1.25em]{\textbf{SE}} \makebox[1.25em]{\textbf{SE}} \makebox[1.25em]{\textbf{AE}} & 50\% & 7.49 \\
        \bottomrule
        \end{tabular} 
    \end{minipage}
\end{table}

\begin{table}[!ht]
    \footnotesize
    \centering
    \caption{
        \textbf{MoE vs. MoE-SEI vs. CDMoE in CEvalBenchmark}.
        % 这些任务来自CEvalBenchmark~\citep{huang2023ceval}, 我们保持Cheems架构中的序列变换部分不变, 使用三种稀疏激活混合专家作为状态变换构造了三个总参数与激活参数几乎相同的模型. 我们详细列出了这三种架构在每个子任务上的零样本和五样本准确率. CDMoE在所有任务上都取得了最好的结果.
        These tasks come from CEvalBenchmark~\citep{huang2023ceval}, we keep the sequence transformation part of the Cheems architecture unchanged, and use three sparse activation mixture of experts as state transformation to construct three models with almost the same total parameters and activation parameters. We detail the zero-shot and five-shot accuracy of these three architectures on each subtask. CDMoE achieved the best results on all tasks.
    }
    \label{tab:comparative_three_moe}
    \begin{tabular}{@{}ccccccccc@{}}
    \toprule
    \sc{Task} & \sc{MoE} & \sc{MoE-SEI} & \sc{CDMoE} & \sc{MoE} & \sc{MoE-SEI} & \sc{CDMoE} \\
    & \sc{zero-shot $\uparrow$} & \sc{zero-shot $\uparrow$} & \sc{zero-shot $\uparrow$} & \sc{five-shot $\uparrow$} & \sc{five-shot $\uparrow$} & \sc{five-shot $\uparrow$} \\
    \midrule
    computer network & 50.11 & 58.74 & 61.07 & 52.95 & 60.23 & 62.36 \\

    operating system & 57.83 & 63.88 & 65.52 & 57.83 & 54.64 & 55.28 \\

    computer architecture & 60.17 & 58.94 & 61.25 & 60.17 & 56.64 & 61.5 \\

    college programming & 37.34 & 38.9 & 42.42 & 37.48 & 38.9 & 41.81 \\

    college physics & 44.2 & 40.12 & 45.98 & 43.7 & 45.98 & 45.98 \\

    college chemistry & 27.17 & 50.74 & 65.03 & 52.73 & 70.15 & 71.5 \\

    advanced mathematics & 42.9 & 44.45 & 48.52 & 45.15 & 48.52 & 48.88 \\

    probability and statistics & 51.82 & 51.82 & 47.8 & 43.32 & 46.77 & 50.9 \\

    discrete mathematics & 38.8 & 45.55 & 41.58 & 38.8 & 40.56 & 42.46 \\

    electrical engineer & 43.86 & 48.17 & 51.53 & 48.17 & 58.12 & 61.99 \\

    metrology engineer & 41.39 & 40.11 & 44.04 & 42.02 & 44.0 & 49.87 \\

    high school mathematics & 44.07 & 46.4 & 47.8 & 44.91 & 46.4 & 48.05 \\

    high school physics & 59.05 & 63.65 & 64.17 & 54.0 & 57.3 & 63.42 \\

    high school chemistry & 24.06 & 55.32 & 57.6 & 24.49 & 55.32 & 57.87 \\

    high school biology & 52.24 & 52.24 & 51.83 & 53.76 & 60.57 & 61.41 \\

    middle school mathematics & 65.93 & 69.27 & 69.65 & 65.38 & 69.27 & 69.99 \\

    middle school biology & 40.65 & 40.95 & 40.65 & 40.65 & 40.95 & 41.68 \\

    middle school physics & 62.47 & 62.47 & 63.5 & 49.38 & 50.88 & 57.16 \\

    middle school chemistry & 44.58 & 46.31 & 53.66 & 44.58 & 44.58 & 52.14 \\

    veterinary medicine & 48.86 & 53.92 & 55.11 & 48.86 & 54.63 & 55.89 \\

    college economics & 26.34 & 39.64 & 36.89 & 29.06 & 35.5 & 39.64 \\

    business administration & 39.88 & 38.68 & 38.68 & 41.67 & 46.28 & 49.14 \\

    marxism & 52.59 & 58.68 & 61.2 & 55.88 & 61.65 & 63.23 \\

    mao zedong thought & 10.91 & 27.62 & 35.47 & 13.39 & 26.99 & 35.47 \\

    education science & 34.38 & 36.18 & 35.41 & 34.75 & 37.14 & 37.14 \\

    teacher qualification & 32.69 & 34.19 & 32.69 & 34.08 & 37.37 & 39.15 \\

    high school politics & 71.84 & 71.13 & 74.56 & 71.84 & 74.94 & 78.4 \\

    high school geography & 55.46 & 55.46 & 55.46 & 55.11 & 55.46 & 56.0 \\

    middle school politics & 35.18 & 38.81 & 39.75 & 32.83 & 38.51 & 41.44 \\

    middle school geography & 75.96 & 78.92 & 76.47 & 49.23 & 50.81 & 66.8 \\

    modern chinese history & 50.31 & 49.34 & 58.36 & 64.35 & 66.02 & 68.61 \\

    ideological and moral cultivation & 27.76 & 51.07 & 53.53 & 30.46 & 47.32 & 51.07 \\

    logic & 61.48 & 61.48 & 61.48 & 59.58 & 61.48 & 60.47 \\

    law & 49.45 & 54.14 & 55.25 & 52.27 & 59.31 & 60.38 \\

    chinese language and literature & 57.36 & 60.78 & 62.24 & 57.04 & 60.78 & 64.17 \\

    art studies & 33.44 & 32.5 & 32.27 & 33.44 & 35.86 & 34.1 \\

    professional tour guide & 50.89 & 52.36 & 52.63 & 48.07 & 48.07 & 51.18 \\

    legal professional & 48.6 & 55.29 & 55.29 & 50.25 & 58.63 & 58.13 \\

    high school chinese & 50.6 & 50.6 & 55.08 & 51.15 & 54.02 & 65.06 \\

    high school history & 33.77 & 47.0 & 47.0 & 43.08 & 57.45 & 59.07 \\

    middle school history & 52.11 & 50.37 & 52.11 & 44.26 & 46.16 & 52.11 \\

    civil servant & 32.97 & 32.65 & 34.83 & 34.34 & 36.9 & 35.07 \\

    sports science & 38.16 & 50.13 & 61.8 & 63.21 & 77.19 & 69.0 \\

    plant protection & 45.81 & 50.45 & 49.17 & 44.73 & 50.45 & 48.47 \\

    basic medicine & 61.37 & 67.15 & 67.15 & 56.81 & 62.71 & 64.27 \\

    clinical medicine & 44.37 & 54.22 & 56.99 & 39.3 & 44.37 & 49.6 \\

    urban and rural planner & 39.01 & 42.95 & 40.43 & 39.54 & 42.95 & 41.06 \\

    accountant & 22.8 & 34.15 & 38.45 & 24.56 & 36.15 & 38.45 \\

    fire engineer & 33.15 & 32.2 & 37.44 & 35.51 & 40.09 & 41.2 \\

    environmental impact assessment & 39.7 & 50.42 & 50.42 & 39.26 & 50.42 & 50.42 \\

    tax accountant & 27.1 & 36.27 & 44.03 & 26.3 & 34.14 & 39.75 \\

    physician & 30.34 & 36.27 & 36.98 & 33.74 & 39.82 & 41.4 \\

    Average & 44.29 & 49.29 & \textbf{51.31} & 44.95 & 50.37 & \textbf{52.88} \\
    \bottomrule
    \end{tabular}
\end{table}

\begin{figure}[ht]
    \centering
    \includegraphics[width=\linewidth]{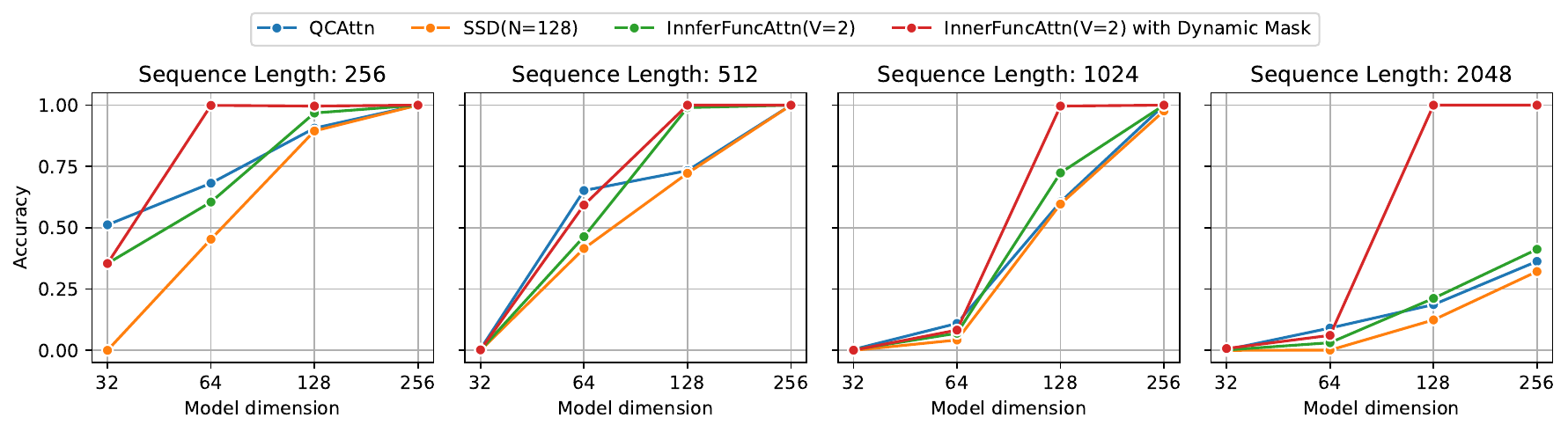}
    \caption{
        \textbf{Multi-Query Associative Recall}.
        % 我们基于原始的多查询联系召回引入了更难的任务版本, 包括更长的序列长度, 更小的模型维度等等, 详细的参数见附录~\ref{sec:evaluation_parameters:multi_query_associative_recall}. InnerFuncAttn with Dynamic Mask 在绝大多数情况下都保持良好的性能.
        We introduced a more difficult version of the original multi-query associative recall task~\citep{arora2024zoology}, including longer sequence lengths, smaller model dimensions, etc. For detailed parameters, see Appendix~\ref{sec:evaluation_parameters:multi_query_associative_recall}. InnerFuncAttn with Dynamic Mask maintains good performance in most cases.
    }
    \label{fig:mqar}
\end{figure}

% 在表~\ref{tab:comparative_pe} 中, 我们可以看到无论是单独使用 SSD 算法, 还是 QCAttn 算法, 或者是两者结合使用, RoPE 在长序列的困惑表现上都是最佳的. 即使是在使用 InnerFuncAttn 和动态掩码的情况下, 也同样如此.
In Table~\ref{tab:comparative_pe}, we can see that whether using the SSD algorithm alone, the QCAttn algorithm, or a combination of the two, RoPE has the best perplexity performance on long sequences. This is also the case when using InnerFuncAttn and dynamic masks.

% 在表~\ref{tab:comparative_ffn} 中, 我们可以看到随着 CDMoE 在整个模型所占的比例增加, 模型在预训练子集上的困惑逐渐降低. 然而我们也发现, 在第一个与最后一个状态变换, 使用完全密集激活的 MLP 模块, 与全部使用 CDMoE 相差不大, 也许这样做可以增加模型输出的稳定性. 但是为了简单建模起见, 我们选择了全部使用 CDMoE.
In Table~\ref{tab:comparative_ffn}, we can see that as the proportion of CDMoE in the entire model increases, the perplexity of the model on the pre-training subset gradually decreases. However, we also found that using a fully dense activation MLP module in the first and last state transformations is not much different from using all CDMoE, which may increase the stability of the model output. However, for the sake of simple modeling, we chose to use all CDMoE.

% 在表~\ref{tab:comparative_three_moe} 中, 我们可以看到在 CEvalBenchmark 中, CDMoE 在所有任务上都取得了最好的结果. 模型参数与表\ref{tab:downstream_evaluation:model}中的1.3B规模参数设置类似. 同时我们也不得不感叹Smollm-Corpus~\citep{benallal2024smollmcorpus}与Chinese Cosmopedia数据集的质量之高, 相较于使用其他训练数据集, 使用它们混合训练, 让这三种模型均提高了10\%左右的准确率, 尤其是在五样本上.
In Table~\ref{tab:comparative_three_moe}, we can see that CDMoE achieved the best results on all tasks in CEvalBenchmark. The model parameters are similar to the 1.3B scale parameter settings in Table~\ref{tab:downstream_evaluation:model}. At the same time, we also have to admire the high quality of the Smollm-Corpus~\citep{benallal2024smollmcorpus} and Chinese Cosmopedia datasets. Compared with using other training datasets, mixing training with them increased the accuracy of these three models by about 10\%, especially in the five-shot.

% 在图~\ref{fig:mqar} 中, 我们展示了 QCAttn, SSD, InnerFuncAttn 和 InnerFuncAttn with Dynamic Mask 在多查询联系召回任务中的性能. 我们可以看到, 当模型维度达到 $128$ 时, 更够为 InnerFuncAttn 每个 $V$ 分配 $64$ 维度的情况下, InnerFuncAttn 是要比 QCAttn 和 SSD 准确率更高的. 而当序列长度延长至 $2048$ 时, 上述三种结构的联系召回能力都被序列噪声所限制, 但 InnerFuncAttn with Dynamic Mask 仍然保持良好的性能.
In Figure~\ref{fig:mqar}, we show the performance of QCAttn, SSD, InnerFuncAttn, and InnerFuncAttn with Dynamic Mask in the multi-query associative recall task. We can see that when the model dimension reaches $128$, and InnerFuncAttn allocates $64$ dimensions for each $V$, InnerFuncAttn is more accurate than QCAttn and SSD. When the sequence length is extended to $2048$, the contact recall capabilities of the three structures are limited by sequence noise, but InnerFuncAttn with Dynamic Mask still maintains good performance.

\subsection{Language Modeling}

\begin{figure}[!ht]
    \centering
    \begin{subfigure}{0.48\linewidth}
        \centering
        \includegraphics[width=\linewidth]{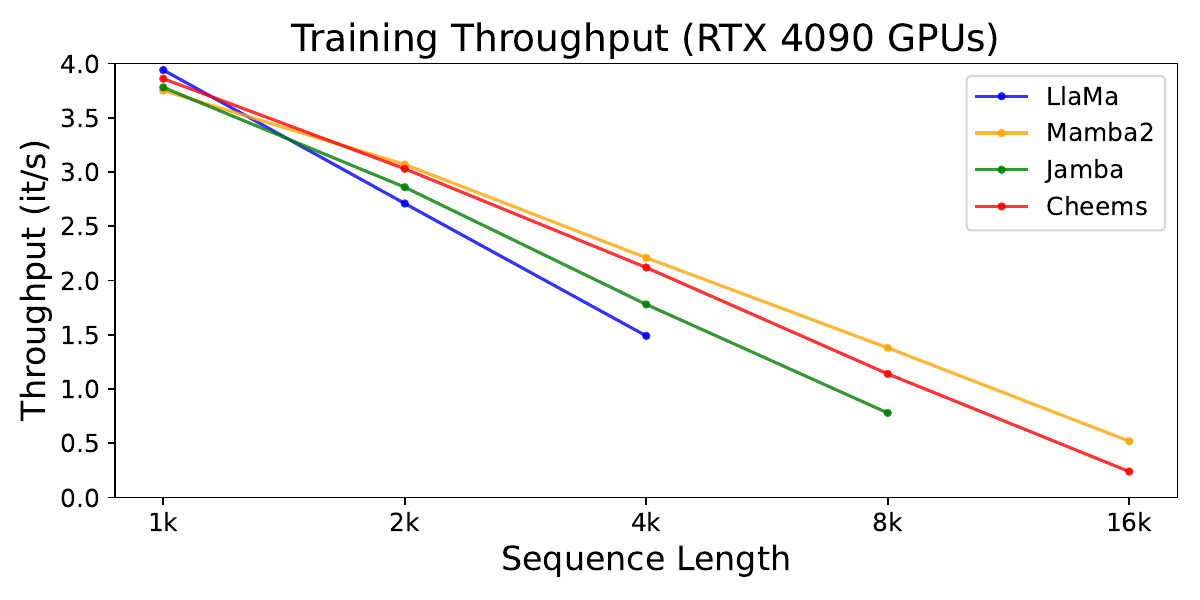}
    \end{subfigure}
    \begin{subfigure}{0.48\linewidth}
        \centering
        \includegraphics[width=\linewidth]{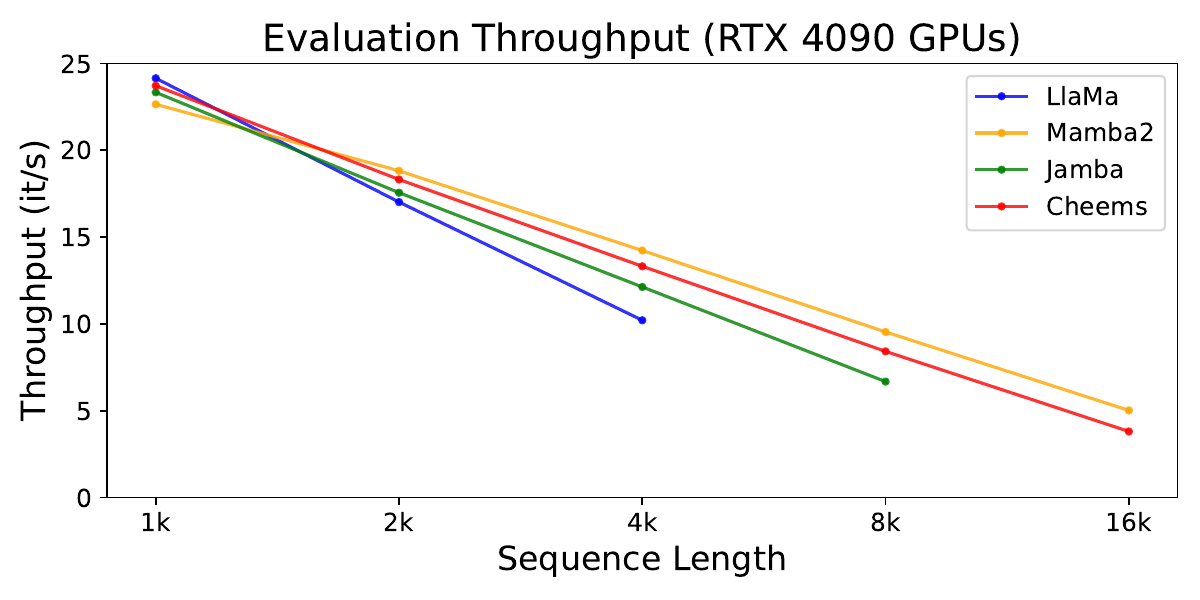}
    \end{subfigure}
    \caption{
        \textbf{Efficient Benchmark}.
        % 使用QCAttn作为序列变换的架构LlaMa, 使用SSD作为序列变换的架构Mamba2, 使用SSD和QCAttn作为序列变换的架构Jamba以及本文提出的Cheems, 这些架构在1.3B参数规模下在不同序列长度进行训练(正向传播和反向传播)与验证(仅正向传播), Cheems的效率超过了LlaMa和Jamba, 但略低于Mamba2. 
        The LlaMa architecture that uses QCAttn as the sequence transformation, the Mamba2 architecture that uses SSD as the sequence transformation, the Jamba architecture that uses SSD and QCAttn as the sequence transformation, and the Cheems architecture proposed in this paper. These architectures are train (both forward and backward) and valid (forward only) at different sequence lengths under the 1.3B parameter scale. Cheems is more efficient than LlaMa and Jamba, but slightly lower than Mamba2.
    }
    \label{fig:efficient_benchmark}
\end{figure}

\begin{table}[!ht]
    \centering
    \caption{
        \textbf{Effective Benchmark}.
        % 使用QCAttn作为序列变换MoE-shared expert isolation作为状态变换的架构LlaMa, 使用SSD作为序列变换MoE-shared expert isolation作为状态变换的架构Mamba2, 使用SSD和QCAttn作为序列变换MoE-shared expert isolation作为状态变换的架构Jamba以及本文提出的Cheems在相同条件下训练的验证结果. 粗体显示每种参数规模的最佳结果, 其次为underline. 对于每种模型参数规模, Cheems的性能大多都优于其他模型. 我们没有给出预训练的困惑表现, 因为模型训练完成时间在transformers库的梯度累计错误修复之前, 对修复前而使用不同梯度累计步数以及修复后的新实验都没有参考价值. 如果今后复现我们的训练结果, 在这些验证指标上的分数可能会有所上升. 验证集介绍与具体模型参数见附录~\ref{sec:evaluation_parameters:downstream_evaluation}.
        The LlaMa architecture that uses QCAttn as the sequence transformation and MoE-SEI as the state transformation, the Mamba2 architecture that uses SSD as the sequence transformation and MoE-SEI as the state transformation, the Jamba architecture that uses SSD and QCAttn as the sequence transformation and MoE-SEI as the state transformation, and the Cheems architecture proposed in this paper. The verification results of the models trained under the same conditions. The best results for each parameter scale are shown in bold, followed by underline. For each model parameter scale, Cheems performs better than other models in most cases. We do not provide the perplexity performance of pre-training because the model training completion time is before the gradient accumulation error fix in the transformers library, and there is no reference value for using different gradient accumulation steps before the fix and new experiments after the fix. If our training results are reproduced in the future, the scores on these verification metrics may rise. For the introduction of the verification set and the specific model parameters, see Appendix~\ref{sec:evaluation_parameters:downstream_evaluation}.
    }
    \label{tab:effective_benchmark}
    \begin{tabular}{@{}ccccccccccccc@{}}
    \toprule
    \sc{Model} & \sc{MMLU} & \sc{TriviaQA} & \sc{ARC} & \sc{PIQA} & \sc{HellaSwag} & \sc{OBQA} & \sc{Winogrande} & \sc{Avg} \\
    & \sc{acc $\uparrow$} & \sc{qem $\uparrow$} & \sc{acc $\uparrow$} & \sc{acc $\uparrow$} & \sc{acc $\uparrow$} & \sc{acc $\uparrow$} & \sc{acc $\uparrow$} & \\
    \midrule
    LlaMa-320M & \underline{33.65} & 8.86 & \textbf{51.68} & 71.42 & 52.30 & \underline{37.02} & 53.15 & 43.99 \\
    Mamba2-320M & 33.10 & \underline{9.36} & 50.72 & 70.24 & 48.62 & 35.16 & 54.17 & 43.07 \\
    Jamba-320M & 33.12 & 9.32 & 50.80 & \underline{71.88} & \underline{52.92} & 36.73 & \underline{55.24} & \underline{44.31} \\
    Cheems-320M & \textbf{34.45} & \textbf{10.38} & \underline{51.57} & \textbf{73.32} & \textbf{53.79} & \textbf{37.42} & \textbf{55.61} & \textbf{45.22} \\
    \midrule
    LlaMa-1.3B & \underline{37.86} & 20.66 & \textbf{59.82} & 76.05 & 61.65 & \textbf{41.15} & 55.40 & 50.36 \\
    Mamba2-1.3B & 36.28 & 21.28 & 58.02 & 72.26 & 59.48 & 37.98 & 58.72 & 49.07 \\
    Jamba-1.3B & 37.43 & \underline{21.60} & 59.33 & \underline{76.58} & \underline{62.33} & 40.82 & \underline{59.20} & \underline{51.07} \\
    Cheems-1.3B & \textbf{39.08} & \textbf{23.02} & \underline{59.69} & \textbf{78.15} & \textbf{63.63} & \underline{41.12} & \textbf{62.09} & \textbf{52.44} \\
    \bottomrule
    \end{tabular}
\end{table}

% 我们选取了分别使用 QCAttn 的 LlaMa 和使用 SSD 算法 Mamba2 以及使用 QCAttn 和 SSD 混合算法的 Jamba 作为 Cheems 的对比对象. 在图~\ref{fig:efficient_benchmark} 中, 我们可以看到Cheems的正反向传播效率已经超过了 LlaMa 和 Jamba, 并且和 Mamba2 保持较低的差距. 在表~\ref{tab:effective_benchmark} 中, 我们可以看到 Cheems 在大多数验证指标上都优于 LlaMa, Mamba2 和 Jamba. 并且随着参数规模的增长, Cheems 的性能提升更为明显.
We selected LlaMa~\citep{touvron2023llama2}, Mamba2~\citep{mamba2}, and Jamba~\citep{lieber2024jamba} as the comparison objects for Cheems.
In Figure~\ref{fig:efficient_benchmark}, we can see that the forward and backward propagation efficiency of Cheems has surpassed LlaMa and Jamba, and maintains a lower gap with Mamba2. In Table~\ref{tab:effective_benchmark}, we can see that Cheems performs better than LlaMa, Mamba2, and Jamba on most verification metrics. And as the parameter scale increases, the performance improvement of Cheems is more significant.

\section{Discussion}
\begin{figure}[!ht]
    \centering
    \includegraphics[height=0.25\textheight]{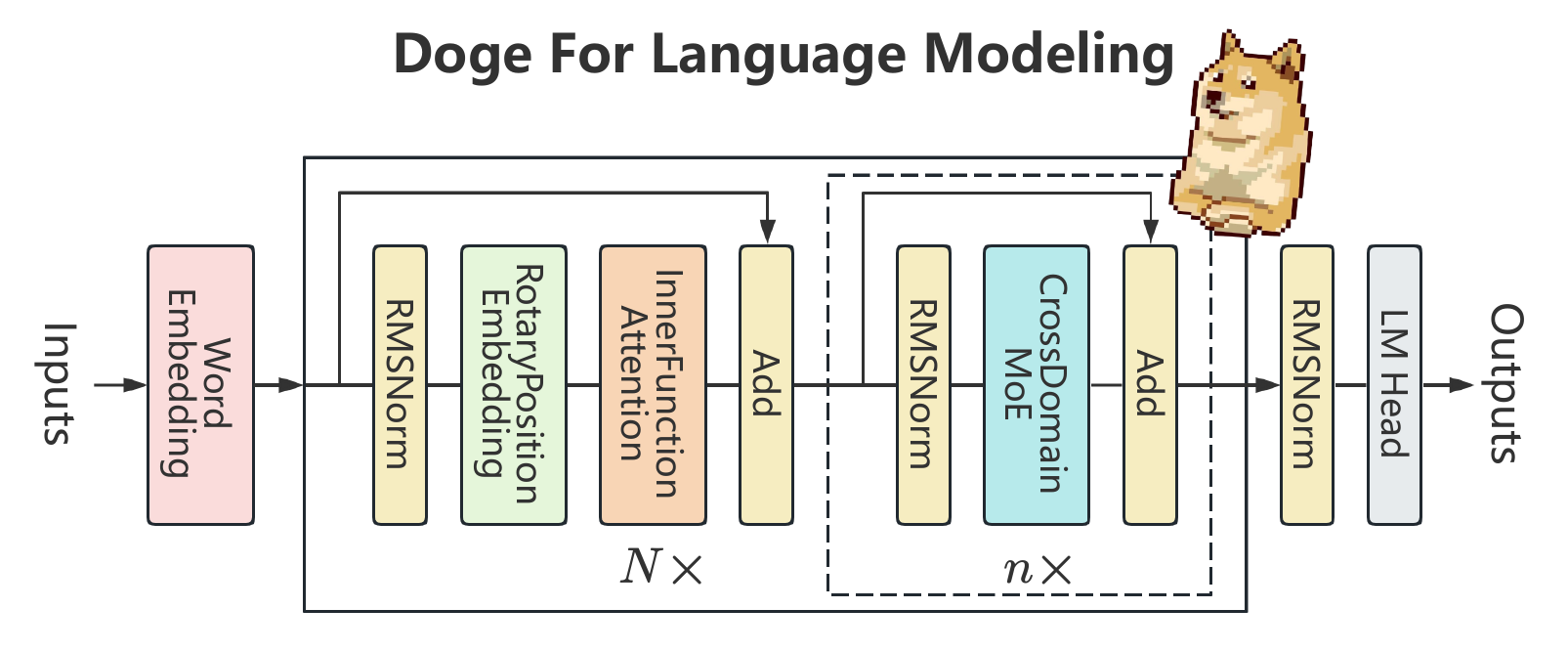}
    \caption{
        \textbf{Doge Architecture}.
        % 将 Cheems 架构中的 SSD 序列变换模块移除, 将架构修改为单个 InnerFuncAttn 序列变换模块之后堆叠多个 CDMoE 这种状态变换模块.
        Remove the SSD sequence transformation module in the Cheems architecture and modify the architecture to stack multiple CDMoE state transformation modules after a single InnerFuncAttn sequence transformation module.
    }
    \label{fig:doge}
\end{figure}

% 实际上, 在我们完成此项工作时遇到了很多问题, 其中包括多种原因导致 mamba-ssm 库无法正常工作. 在这个问题解决之前, 我们尝试了直接将 SSD 序列变换模块移除, 将架构修改为如图~\ref{fig:doge}, 单个 Attn 序列变换模块之后堆叠多个 MLP 或者 MoE 这种状态变换模块. 我们发现使用这种模型架构, 保证参数量与Transformer架构相等或者更少的情况下进行语言建模, 在绝大多数验证指标上都没有明显下降. 我们猜测目前的Transformer架构中, Attn 层是可能存在部分冗余的. 研究注意力分数的模型层影响深度, 减少注意力冗余, 可能会是未来的研究方向.
In fact, we encountered many problems when completing this work, including various reasons that caused the mamba-ssm library to not work properly. Before solving this problem, we tried to directly remove the SSD sequence transformation module and modify the architecture to stack multiple MLP or MoE state transformation modules after a single Attn sequence transformation module, as shown in Figure~\ref{fig:doge}. We found that using this model architecture, ensuring that the number of parameters is equal to or less than the Transformer architecture for language modeling, there is no significant decrease in most verification metrics. We speculate that in the current Transformer architecture, the Attn layer may have some redundancy. Research on the impact of model layer attention scores on depth and reducing attention redundancy may be a future research direction.

\section{Conclusion}
\label{sec:conclusion}

% 本文探讨了将状态空间双算法与二次因果自注意力算法融合进行建模的思路. 研究了在融合算法下的高效位置编码, 可以增强表达能力的内函数注意力 with dynamic mask和减少参数冗余的交叉领域混合专家, 最后验证了这些算法在语言建模的性能均达到先进水平, 推动语言建模朝着更加高效和有效的方向发展.
This paper explores the idea of modeling by integrating the state space duality algorithm with the quadratic causal self-attention algorithm. We studied the efficient positional encoding under the fusion algorithm, the internal function attention with dynamic mask that can enhance the expressive power, and the cross domain mixture of experts that reduces parameter redundancy. Finally, we verified that these algorithms have reached the advanced level in language modeling performance, promoting the development of language modeling in a more efficient and effective direction.

\subsubsection*{Acknowledgments}
% 我们感谢我们的家庭, 对我们作为独立研究员完成这项工作的理解与支持. 同时, 我们也感谢卡内基梅隆大学的Albert Gu教授, 为我们提供了ArXiv的背书, 使我们能够在本科阶段投身于科学研究.
We thank our families for their understanding and support in completing this work as independent researchers. At the same time, we also thank Professor Albert Gu of Carnegie Mellon University for providing us with an endorsement of ArXiv, allowing us to engage in scientific research at the undergraduate level.

\printbibliography

\newpage

\appendix

\onecolumn
  
\section{RoPE for SSD}
\label{sec:rope_for_ssd}

\begin{proof}[Proof of equation~\ref{eq:rope_cb}]
by definition, $h_0 = B_0 x_0$.
By induction,

\begin{align*}
    h_t &= A_t \dots A_1 B_0 x_0 + A_t \dots A_2 B_1 x_1 + \dots + A_t A_{t-1} B_{t-2} x_{t-2} + A_t B_{t-1} x_{t-1} + B_t x_t
    \\&= \sum_{s=0}^t A_{t:s}^\times B_s x_s
\end{align*}

Multiplying by $C_t$ to produce $y_t$, and vectorizing the equation to $t \in [\mathtt{T}]$ ($\mathtt{T}$ is the sequence length), we derive the matrix transformation form of SSD.

\begin{align*}
    y_t &= \sum_{s=0}^t C_t^{\top} A_{t:s}^\times B_s x_s
    \\
    y &= \mathsf{SSD}(A, B, C)(x) = Mx
    \\
    M_{ji} &\coloneqq C_j^{\top} A_{j} \cdots A_{i+1} B_{i}
\end{align*}

Then the matrix form of SSD is represented using SSS (Sequentially Semiseparable) as $M = \mathsf{SSS}(A, B, C)$, where $M_{ji} = C_j^{\top} A_{j:i} B_i$, and then considering $A$ is just a scalar, rearranged as

\begin{align*}
    M_{ji} = A_{j:i} \cdot (C_j^{\top}B_i)
\end{align*}

Vectorized as

\begin{align*}
    L &\coloneqq \mathsf{1SS}(a) \\
    M &= L \circ (C B^{\top}) \\
\end{align*}

Finally, it is proved that the matrix transformation form of SSD is equivalent to Attention $(L \circ QK^\top) \cdot V = (L \circ CB^\top) \cdot X$.

Now we have enough theoretical support to give rotational positional encoding to the $C$ and $B$ matrices in SSD.

\begin{subequations}
    \begin{align*}
        C_{j} &= f_{C} (x_{j}, j) \\
        B_{i} &= f_{B} (x_{i}, i)
    \end{align*}
\end{subequations}

$C_{j}$ represents the output weight matrix of the $j$-th token corresponding to the word vector $x_{j}$ integrated with the position information $j$,
$B_{i}$ represents the input weight matrix of the $i$-th token corresponding to the word vector $x_{i}$ integrated with the position information $i$.

To utilize the relative positional information between tokens,
we assume that the inner product operation between the $C_{j}$ vector and the $B_{i}$ vector can be represented by a function $g$,
where the input of the function $g$ is the word embedding vectors $x_{j}$ and $x_{i}$,
and their relative positional information $j - i$,
the inner product of $C_{j}$ and $B_{i}$ and their relative positional information $j - i$ is defined as

\begin{equation*}
    <f_{C}(x_{j}, j), f_{B}(x_{i}, i)> = g(x_{j}, x_{i}, j - i)
\end{equation*}

Now, assuming the word embedding vector dimension is $d = 2$,
we have $f_{C}(x_{j}, j) = (W_{C} x_{j})e^{\imath j \theta}$,
for the first half of the formula $W_{C} x_{j}$,
we know that $W_{C}$ is a two-dimensional matrix,
$x_{j}$ is a two-dimensional vector,
the result of the multiplication is naturally a two-dimensional vector,
represented by $C_{j}$

\begin{align*}
    C_{j} &= 
        \begin{bmatrix} 
            C_{j}^{(1)} \\
            C_{j}^{(2)}
        \end{bmatrix}
    = W_{C} x_{j} = 
        \begin{bmatrix} 
            W_{C}^{(11)} & W_{C}^{(12)} \\
            W_{C}^{(21)} & W_{C}^{(22)}
        \end{bmatrix}
        \begin{bmatrix} 
            x_{j}^{(1)} \\
            x_{j}^{(2)}
        \end{bmatrix}
\end{align*}

For the second half $e^{\imath j \theta}$,
according to Euler's formula $e^{\imath x} = \cos(x) + \imath \sin(x)$,
we have

\begin{align*}
    e^{\imath j \theta} &= \cos(j \theta) + \imath \sin(j \theta)
\end{align*}

We know

\begin{align*}
    f_{C}(x_{j}, j) &= (W_{C} x_{j})e^{\imath j \theta} = C_{j}e^{\imath j \theta}
\end{align*}

$C_{j}$ is represented in complex form,

\begin{align*}
    C_{j} &=
        \begin{bmatrix} 
            C_{j}^{(1)}, C_{j}^{(2)}
        \end{bmatrix}
    = 
        \begin{bmatrix} 
            C_{j}^{(1)} + \imath C_{j}^{(2)}
        \end{bmatrix}
\end{align*}

Thus,

\begin{align*}
    f_{C}(x_{j}, j) &= C_{j}e^{\imath j \theta} = 
        \begin{bmatrix} 
            C_{j}^{(1)} + \imath C_{j}^{(2)}
        \end{bmatrix} e^{\imath j \theta}
\end{align*}

According to the above derivation,
we know that $f_{C}(x_{j}, j)$ is the product of two complex numbers,

\begin{align*}
    f_{C}(x_{j}, j) &= C_{j}e^{\imath j \theta} = 
        \begin{bmatrix} 
            C_{j}^{(1)} + \imath C_{j}^{(2)}
        \end{bmatrix} \times (\cos(j \theta) + \imath \sin(j \theta))
\end{align*}

Considering the following two formulas about complex numbers

\begin{align*}
    (a + \imath b) \times (c + \imath d) &= ac + \imath bc + \imath ad + \imath^2 bd = (ac - bd) + \imath (bc + ad) \\
    \imath^2 &= -1
\end{align*}

We have

\begin{align*}
    C_{j}e^{\imath j \theta} &= 
        \begin{bmatrix} 
            C_{j}^{(1)} + \imath C_{j}^{(2)}
        \end{bmatrix} \times (\cos(j \theta) + \imath \sin(j \theta)) = 
        \begin{bmatrix} 
            C_{j}^{(1)} \cos(j \theta) - C_{j}^{(2)} \sin(j \theta)
        \end{bmatrix} + \imath
        \begin{bmatrix} 
            C_{j}^{(2)} \cos(j \theta) + C_{j}^{(1)} \sin(j \theta)
        \end{bmatrix}
\end{align*}

Expressing this result as a real vector,

\begin{align*}
    C_{j}e^{\imath j \theta} &= 
        \begin{bmatrix} 
            C_{j}^{(1)} \cos(j \theta) - C_{j}^{(2)} \sin(j \theta),
            C_{j}^{(2)} \cos(j \theta) + C_{j}^{(1)} \sin(j \theta)
        \end{bmatrix}
\end{align*}

Therefore, $C_{j}$ multiplied by a rotation matrix is obtained.

\begin{align*}
    f_{C}(x_{j}, j) &= (W_{C} x_{j})e^{\imath j \theta} = C_{j}e^{\imath j \theta} \\
    &=
    \begin{bmatrix} 
        C_{j}^{(1)} \cos(j \theta) - C_{j}^{(2)} \sin(j \theta),
        C_{j}^{(2)} \cos(j \theta) + C_{j}^{(1)} \sin(j \theta)
    \end{bmatrix} \\
    &=
    \begin{bmatrix} 
        \cos(j \theta) & -\sin(j \theta) \\
        \sin(j \theta) & \cos(j \theta)
    \end{bmatrix}
    \begin{bmatrix} 
        C_{j}^{(1)} \\
        C_{j}^{(2)}
    \end{bmatrix}
\end{align*}

Similarly, $B_{i}$ vector can be obtained

\begin{align*}
    f_{B}(x_{i}, i) &= (W_{B} x_{i})e^{\imath i \theta} = B_{i}e^{\imath i \theta} \\
    &=
    \begin{bmatrix} 
        B_{i}^{(1)} \cos(i \theta) - B_{i}^{(2)} \sin(i \theta),
        B_{i}^{(2)} \cos(i \theta) + B_{i}^{(1)} \sin(i \theta)
    \end{bmatrix} \\
    &=
    \begin{bmatrix} 
        \cos(i \theta) & -\sin(i \theta) \\
        \sin(i \theta) & \cos(i \theta)
    \end{bmatrix}
    \begin{bmatrix} 
        B_{i}^{(1)} \\
        B_{i}^{(2)}
    \end{bmatrix}
\end{align*}

The function $g$ can be represented as

\begin{align*}
    g(x_{j}, x_{i}, j - i) &= \Re 
        \begin{bmatrix} 
            (W_{C} x_{j})(W_{B} x_{i})^*e^{\imath (j - i) \theta}
        \end{bmatrix}
\end{align*}

where $\Re$ represents the real part of the complex number $x$,
$(W_{C} x_{j})(W_{B} x_{i})^*$ represents the conjugate of the product of two complex numbers.
Considering

\begin{align*}
    z &= a + \imath b \\
    z^* &= a - \imath b
\end{align*}

we have

\begin{align*}
    W_{C} x_{j} &= C_{j} = C_{j}^{(1)} + \imath C_{j}^{(2)} \\
    W_{B} x_{i} &= B_{i} = B_{i}^{(1)} + \imath B_{i}^{(2)} \\
    (W_{B} x_{i})^* &= B_{i}^* = B_{i}^{(1)} - \imath B_{i}^{(2)} \\
    e^{\imath (j - i) \theta} &= \cos((j - i) \theta) + \imath \sin((j - i) \theta)
\end{align*}

We now want to prove that

\begin{align*}
    g(x_{j}, x_{i}, j - i) &= \Re 
        \begin{bmatrix} 
            (W_{C} x_{j})(W_{B} x_{i})^*e^{\imath (j - i) \theta}
        \end{bmatrix} \\
    &= \Re
        \begin{bmatrix} 
            (C_{j}^{(1)} + \imath C_{j}^{(2)})(B_{i}^{(1)} - \imath B_{i}^{(2)})(\cos((j - i) \theta) + \imath \sin((j - i) \theta))
        \end{bmatrix} \\
    &= \Re
        \begin{bmatrix} 
            ((C_{j}^{(1)}B_{i}^{(1)} + C_{j}^{(2)}B_{i}^{(2)}) + \imath (C_{j}^{(2)}B_{i}^{(1)} - C_{j}^{(1)}B_{i}^{(2)}))(\cos((j - i) \theta) + \imath \sin((j - i) \theta))
        \end{bmatrix} \\
    &= (C_{j}^{(1)}B_{i}^{(1)} + C_{j}^{(2)}B_{i}^{(2)})\cos((j - i) \theta) - (C_{j}^{(2)}B_{i}^{(1)} - C_{j}^{(1)}B_{i}^{(2)})\sin((j - i) \theta)
\end{align*}

Recalling the vectorized form of SSD,
the $C$ vector at position $j$ and the $B$ vector at position $i$ will perform an inner product operation,
that is,

\begin{align*}
    f_{C}(x_{j}, j) &= 
        \begin{bmatrix} 
            C_{j}^{(1)} \cos(j \theta) - C_{j}^{(2)} \sin(j \theta),
            C_{j}^{(2)} \cos(j \theta) + C_{j}^{(1)} \sin(j \theta)
        \end{bmatrix} \\
    f_{B}(x_{i}, i) &=
        \begin{bmatrix} 
            B_{i}^{(1)} \cos(i \theta) - B_{i}^{(2)} \sin(i \theta),
            B_{i}^{(2)} \cos(i \theta) + B_{i}^{(1)} \sin(i \theta)
        \end{bmatrix}
\end{align*}

We have

\begin{align*}
    <f_{C}(x_{j}, j), f_{B}(x_{i}, i)> &= 
        \begin{bmatrix} 
            C_{j}^{(1)} \cos(j \theta) - C_{j}^{(2)} \sin(j \theta)
        \end{bmatrix}
        \begin{bmatrix} 
            B_{i}^{(1)} \cos(i \theta) - B_{i}^{(2)} \sin(i \theta)
        \end{bmatrix} \\
        &+
        \begin{bmatrix} 
            C_{j}^{(2)} \cos(j \theta) + C_{j}^{(1)} \sin(j \theta)
        \end{bmatrix}
        \begin{bmatrix} 
            B_{i}^{(2)} \cos(i \theta) + B_{i}^{(1)} \sin(i \theta)
        \end{bmatrix} \\
        &= C_{j}^{(1)} \cos(j \theta) B_{i}^{(1)} \cos(i \theta) - C_{j}^{(1)} \cos(j \theta) B_{i}^{(2)} \sin(i \theta) \\
        &- C_{j}^{(2)} \sin(j \theta) B_{i}^{(1)} \cos(i \theta) + C_{j}^{(2)} \sin(j \theta) B_{i}^{(2)} \sin(i \theta) \\
        &+ C_{j}^{(2)} \cos(j \theta) B_{i}^{(2)} \cos(i \theta) + C_{j}^{(2)} \cos(j \theta) B_{i}^{(1)} \sin(i \theta) \\
        &+ C_{j}^{(1)} \sin(j \theta) B_{i}^{(2)} \cos(i \theta) + C_{j}^{(1)} \sin(j \theta) B_{i}^{(1)} \sin(i \theta)
\end{align*}

Considering

\begin{align*}
    \sin(a + b) &= \sin(a)\cos(b) + \cos(a)\sin(b) \\
    \sin(a - b) &= \sin(a)\cos(b) - \cos(a)\sin(b) \\
    \cos(a + b) &= \cos(a)\cos(b) - \sin(a)\sin(b) \\
    \cos(a - b) &= \cos(a)\cos(b) + \sin(a)\sin(b)
\end{align*}

We have

\begin{align*}
    <f_{C}(x_{j}, j), f_{B}(x_{i}, i)> &= 
        C_{j}^{(1)} B_{i}^{(1)} (\cos(j \theta) \cos(i \theta) + \sin(j \theta) \sin(i \theta)) \\
        &+ C_{j}^{(1)} B_{i}^{(2)} (-\cos(j \theta) \sin(i \theta) + \sin(j \theta) \cos(i \theta)) \\
        &+ C_{j}^{(2)} B_{i}^{(1)} (-\sin(j \theta) \cos(i \theta) + \cos(j \theta) \sin(i \theta)) \\
        &+ C_{j}^{(2)} B_{i}^{(2)} (\sin(j \theta) \sin(i \theta) + \cos(j \theta) \cos(i \theta)) \\
        &= C_{j}^{(1)} B_{i}^{(1)} \cos((j - i) \theta) + C_{j}^{(1)} B_{i}^{(2)} \sin((j - i) \theta) \\
        &- C_{j}^{(2)} B_{i}^{(1)} \sin((j - i) \theta) + C_{j}^{(2)} B_{i}^{(2)} \cos((j - i) \theta) \\
        &= (C_{j}^{(1)} B_{i}^{(1)} + C_{j}^{(2)} B_{i}^{(2)})\cos((j - i) \theta) + (C_{j}^{(1)} B_{i}^{(2)} - C_{j}^{(2)} B_{i}^{(1)})\sin((j - i) \theta) \\
        &= (C_{j}^{(1)}B_{i}^{(1)} + C_{j}^{(2)}B_{i}^{(2)})\cos((j - i) \theta) - (C_{j}^{(2)}B_{i}^{(1)} - C_{j}^{(1)}B_{i}^{(2)})\sin((j - i) \theta) \\
        &= g(x_{j}, x_{i}, j - i)
\end{align*}

It is proved that the inner product of the $C$ vector at position $j$ and the $B$ vector at position $i$ is the function $g$.

Finally, using the matrix-vector multiplication form

\begin{align*}
    <f_{C}(x_{j}, j), f_{B}(x_{i}, i)> &= 
        \begin{bmatrix} 
            \begin{bmatrix} 
                \cos(j \theta) & -\sin(j \theta) \\
                \sin(j \theta) & \cos(j \theta)
            \end{bmatrix}
            \begin{bmatrix} 
                C_{j}^{(1)} \\
                C_{j}^{(2)}
            \end{bmatrix}
        \end{bmatrix}^{T}
        \begin{bmatrix} 
            \begin{bmatrix} 
                \cos(i \theta) & -\sin(i \theta) \\
                \sin(i \theta) & \cos(i \theta)
            \end{bmatrix}
            \begin{bmatrix} 
                B_{i}^{(1)} \\
                B_{i}^{(2)}
            \end{bmatrix}
        \end{bmatrix} \\
        &= 
        \begin{bmatrix} 
            C_{j}^{(1)} & C_{j}^{(2)}
        \end{bmatrix}
        \begin{bmatrix} 
            \cos(j \theta) & \sin(j \theta) \\
            -\sin(j \theta) & \cos(j \theta)
        \end{bmatrix}
        \begin{bmatrix} 
            \cos(i \theta) & -\sin(i \theta) \\
            \sin(i \theta) & \cos(i \theta)
        \end{bmatrix}
        \begin{bmatrix} 
            B_{i}^{(1)} \\
            B_{i}^{(2)}
        \end{bmatrix} \\
\end{align*}

Expanding the product of the two rotary matrices, we have

\begin{align*}
    \begin{bmatrix} 
        \cos(j \theta) \cos(i \theta) + \sin(j \theta) \sin(i \theta) & -\cos(j \theta) \sin(i \theta) + \sin(j \theta) \cos(i \theta) \\
        -\sin(j \theta) \cos(i \theta) + \cos(j \theta) \sin(i \theta) & \sin(j \theta) \sin(i \theta) + \cos(j \theta) \cos(i \theta)
    \end{bmatrix}
\end{align*}

Finally, we get

\begin{align*}
    <f_{C}(x_{j}, j), f_{B}(x_{i}, i)> &= 
        \begin{bmatrix} 
            C_{j}^{(1)} & C_{j}^{(2)}
        \end{bmatrix}
        \begin{bmatrix} 
            \cos((j - i) \theta) & -\sin((j - i) \theta) \\
            \sin((j - i) \theta) & \cos((j - i) \theta)
        \end{bmatrix}
        \begin{bmatrix} 
            B_{i}^{(1)} \\
            B_{i}^{(2)}
        \end{bmatrix}
\end{align*}

The above derivation is only for the case of word embedding dimension $d = 2$,
when $d > 2$, the two-dimensional case can be extended to any dimension as follows

\begin{align*}
    f_{\{C, B\}}(x_{j}, j) &= \mathbb{R}_{\Theta, j}^{d} W_{\{C, B\}} x_{j}
\end{align*}

The inner product satisfies linearity,
so for any even-dimensional RoPE, we can represent it as a concatenation of the two-dimensional case,
that is, grouping the elements of the word embedding vector in pairs

\begin{align*}
    \mathbb{R}_{\Theta, j}^{d} = \begin{bmatrix} 
        \cos j \theta_0 & -sin j \theta_0 & 0 & 0 & \dots & 0 & 0 \\
        \sin j \theta_0 & \cos j \theta_0 & 0 & 0 & \dots & 0 & 0 \\
        0 & 0 & \cos j \theta_1 & -sin j \theta_1 & \dots & 0 & 0 \\
        0 & 0 & \sin j \theta_1 & \cos j \theta_1 & \dots & 0 & 0 \\
        \vdots & \vdots & \vdots & \vdots & \ddots & \vdots & \vdots \\
        0 & 0 & 0 & 0 & \dots & \cos j \theta_{d/2} & -sin j \theta_{d/2-1} \\
        0 & 0 & 0 & 0 & \dots & \sin j \theta_{d/2} & \cos j \theta_{d/2-1} \\
    \end{bmatrix}
\end{align*}

Each group applies the same rotation operation and the rotation angle of each group is calculated as follows:

\begin{align*}
    \Theta &= \{\theta_i = 10000^{-2(i - 1) / d}, i \in [1, 2, \dots, d / 2]\}
\end{align*}
  
% RoPE is a kind of relative positional encoding, and relative positional encoding is a special form of Toeplitz matrix

% \begin{align*}
%     \begin{bmatrix} 
%         0 & & & & & & \\
%         1 & 0 & & & & \\
%         2 & 1 & 0 & & \\
%         3 & 2 & 1 & 0 \\
%         \vdots & \ddots & \ddots & \ddots & \ddots \\
%         n-1 & n-2 & n-3 & \dots & 1 & 0
%     \end{bmatrix}
% \end{align*}

% We can know that the position distribution after RoPE is unbalanced, 0 appears most frequently as the low bit, and n-1 appears least frequently as the high bit, which also leads to a problem of RoPE, its high bit is not sufficiently trained, and the generalization ability is not as good as the low bit. We take the average value of the effective sequence length of the training data as the base, for the sequence length greater than the base, we have

% \begin{align*}
%     C \times \max(1, \log_{base} n) \\
% \end{align*}

% The part of the sequence length within the base is not affected, and the part of the sequence length greater than the base is expanded according to the ratio of $\log_{base} n$, so that the problem of insufficient generalization ability of the high bit can be solved.

\end{proof}

\newpage
\section{Implementation Code}
\label{sec:implementation_code}

\subsection{RoPE}
\label{sec:implementation_code:rope}

\begin{listing}[ht]
    \begin{lstlisting}[language=Python,breaklines]
class RotaryEmbedding:
    def __init__(self,dim,max_position_embeddings,base=10000,scaling_factor=1.0):
        self.dim,self.base,self.max_position_embeddings,self.scaling_factor=dim,base,max_position_embeddings,scaling_factor
        inv_freq=1.0/(self.base**(torch.arange(0,self.dim,2)/self.dim))
        self.register_buffer("inv_freq",inv_freq)

    def forward(self,x,position_ids):
        seq_len=torch.max(position_ids)+1
        if seq_len > self.max_position_embeddings:
            base=self.base*((self.scaling_factor*seq_len/self.max_position_embeddings)-(self.scaling_factor-1))**(self.dim/(self.dim-2))
            inv_freq=1.0/(base**(torch.arange(0,self.dim,2)/self.dim))
        else:
            inv_freq = self.inv_freq
        inv_freq_expanded=inv_freq[None,:,None].expand(position_ids.shape[0],-1,1)
        position_ids_expanded=position_ids[:,None,:]
        freqs=(inv_freq_expanded@position_ids_expanded).transpose(1,2)
        emb=torch.cat((freqs,freqs),dim=-1)
        cos,sin=emb.cos().to(x.dtype),emb.sin().to(x.dtype)
        return cos,sin

def rotate_half(x):
    x1,x2 = x[..., : x.shape[-1] // 2],x[..., x.shape[-1] // 2 :]
    return torch.cat((-x2, x1), dim=-1)

def apply_QK_rotary_pos_emb(q,k,cos,sin,unsqueeze_dim=2):
    cos,sin = cos.unsqueeze(unsqueeze_dim),sin.unsqueeze(unsqueeze_dim)
    q_embed=(q*cos)+(rotate_half(q)*sin)
    k_embed=(k*cos)+(rotate_half(k)*sin)
    return q_embed,k_embed

def apply_CB_rotary_pos_emb(c,b,cos,sin,unsqueeze_dim=2):
    cos,sin = cos.unsqueeze(unsqueeze_dim),sin.unsqueeze(unsqueeze_dim)
    c_embed=(c*cos)+(rotate_half(c)*sin)
    b_embed=(b*cos)+(rotate_half(b)*sin)
    return c_embed,b_embed
    \end{lstlisting}
    \caption{PyTorch example of RoPE.}
    \label{lst:rope}
\end{listing}
\clearpage

\subsection{SSD}
\label{sec:implementation_code:ssd}

\begin{listing}[ht]
    \begin{lstlisting}[language=Python,breaklines]
def pad_tensor_by_size(input_tensor, pad_size):
    # Pad seq_len to be multiple of chunk_len
    return F.pad(input_tensor,(0,0,0,0,0,pad_size,0,0) if len(input_tensor.shape)==4 else (0,0,0,pad_size,0,0))

def reshape_into_chunks(input_tensor,pad_size,chunk_len):
    # Padding input_tensor with `pad_size` on the seq_len dim (dim=1) and
    simultaneously splitting it into chunk sequences.
    # b t ... -> b (l c) ...
    if len(pad_tensor_by_size(input_tensor,pad_size).shape)==3:
        return rearrange(input_tensor,'b(lc)h->blch',c=chunk_len)
    else:
        return rearrange(input_tensor,'b(lc)hd->blchd',c=chunk_len)

def segment_sum(input_tensor):
    # More stable segment sum calculation. Uses cumulative sums and masking instead of direct subtractions.
    chunk_len=input_tensor.size(-1)
    # 1. expand input tensor to have an additional dimension and repeat along that dimension
    # [..., chunk_len] -> [..., chunk_len, chunk_len]
    input_tensor=input_tensor[...,None].expand(*input_tensor.size(),chunk_len)
    # 2. create a lower triangular mask with the diagonal set to 0 to 0 out elements above diag
    mask=torch.tril(torch.ones(chunk_len,chunk_len,device=input_tensor.device,dtype=torch.bool),diagonal=-1)
    input_tensor=input_tensor.masked_fill(~mask,0)
    # 3. compute actual cumsum
    tensor_segsum=torch.cumsum(input_tensor,dim=-2)
    # 4. apply mask to keep only the lower triangular part of the cumulative sum result (incl diagonal this time)
    mask=torch.tril(torch.ones(chunk_len,chunk_len,device=input_tensor.device,dtype=torch.bool),diagonal=0)
    tensor_segsum=tensor_segsum.masked_fill(~mask,-torch.inf)
    return tensor_segsum
    \end{lstlisting}
    \caption{Example SSD helper function in PyTorch}
    \label{lst:ssd_helper}
\end{listing}

\begin{listing}[ht]
    \begin{lstlisting}[language=Python,breaklines]
def ssd(x,dt,A,B,C,chunk_len,D):
    seq_len = x.size(1)
    pad_size=(chunk_len-seq_len%chunk_len)%chunk_len
    D_residual=rearrange(D,'...->...1')*pad_tensor_by_size(x, pad_size)
    # Discretize x and A
    x,A=x*rearrange(dt,'...->...1'),A.to(x.dtype)*dt
    # Rearrange into blocks/chunks
    x,A,B,C=[reshape_into_chunks(t,pad_size,chunk_len) for t in (x,A,B,C)]
    # Compute cumulative sum of A
    A=rearrange(A,'bclh->bhcl',l=chunk_len)
    A_cumsum=torch.cumsum(A,dim=-1)
    # 1. Compute the output for each intra-chunk (diagonal blocks)
    # This is the analog of a causal mask
    L=torch.exp(segment_sum(A))
    # First, contraction of C and B to get G (attention-weights like)
    G=(rearrange(C,'blchn->blc1hn')*rearrange(B,'blchn->bl1chn')).sum(dim=-1) # shape: (b, c, l, s, h)
    # Step 2: Compute M, equivalent to applying attention mask to weights
    M_intermediate=rearrange(G,'...->...1')*rearrange(L,'bhcst->bcsth1')
    M=M_intermediate.sum(dim=-1)
    # Step 3: Compute Y_diag (apply to values)
    Y_diag=(rearrange(M,'...->...1')*rearrange(x,'blchp->bl1chp')).sum(3)
    # (right term of low-rank factorization of off-diagonal blocks; B terms)
    decay_states=torch.exp((A_cumsum[:,:,:,-1:]-A_cumsum))
    B_decay_contraction=B*rearrange(decay_states,'bhcl->bclh1')
    # permute back B * decay states
    states=(rearrange(B_decay_contraction,'bclhs->bchls1')*rearrange(x,'blchp->blhc1p')).sum(dim=3).permute(0,1,2,4,3)
    previous_states=torch.zeros_like(states[:,:1])
    states=torch.cat([previous_states,states],dim=1)
    decay_chunk=torch.exp(segment_sum(nn.functional.pad(A_cumsum[:,:,:,-1],(1,0))))
    states_permuted=states.permute(0,2,1,3,4)
    result=(decay_chunk[...,None,None]*states_permuted[:,:,None,...]).sum(dim=2)
    new_states=result.permute(0, 2, 1, 3, 4)
    states=new_states[:, :-1]
    # Compute state -> output conversion per chunk
    # (left term of low-rank factorization of off-diagonal blocks; C terms)
    # compute Yoff
    C_times_states=rearrange(C,'bclhn->bclh1n')*rearrange(states,'bchpn->bc1hpn')
    Y_off=(C_times_states.sum(-1)*rearrange(torch.exp(A_cumsum),'bhcl->bclh1'))
    # Add output of intra-chunk and inter-chunk terms (diagonal and off-diagonal blocks)
    y=rearrange(Y_diag+Y_off,'bclhp->b(cl)hp')+D_residual
    # Cutting off padded chunks
    if pad_size > 0:
        y=y[:,:seq_len,:,:]
    \end{lstlisting}
    \caption{Example SSD algorithm in PyTorch. We have changed some slow methods to faster ones. The original SSD algorithm implementation can be found in the Mamba2 paper.}
    \label{lst:ssd_algorithm_pytorch}
\end{listing}

\begin{listing}[ht]
    \begin{lstlisting}[language=Python,breaklines]
class SSD(nn.Module):
    def __init__(self,d_model,n_heads,d_head,n_groups,d_state,chunk_len):
    super().__init__()
    self.n_heads,self.n_groups,self.d_state,self.chunk_len=n_heads,n_groups,d_state,chunk_len
    # Initialize parameters
    self.X_proj=nn.Linear(d_model,self.n_heads*d_head)
    self.A_log=nn.Parameter(torch.log(torch.arange(1,self.n_heads+1)))
    self.B_proj=nn.Linear(d_model,self.n_groups*self.d_state)
    self.C_proj=nn.Linear(d_model,self.n_groups*self.d_state)
    self.dt_proj=nn.Linear(d_model,self.n_heads)
    self.D=nn.Parameter(torch.ones(self.n_heads))
    self.out_proj=nn.Linear(d_model,d_model)

    def forward(self, x):
        """
        Notations: b - batch size d - d_model 
                    h - n_heads p - d_head n - d_state g - n_groups
                    t - target sequence length s - source sequence length
                    c - n_chunks l - chunk_len
        """
        dtype=x.dtype
        A=-torch.exp(self.A_log.float())
        B=rearrange(self.B_proj(x),'bt(gn)->btgn',g=self.n_groups,n=self.d_state).repeat(1,1,self.n_heads//self.n_groups,1)
        C=rearrange(self.C_proj(x),'bt(gn)->btgn',g=self.n_groups,n=self.d_state).repeat(1,1,self.n_heads//self.n_groups,1)
        dt=self.dt_proj(x)
        x=rearrange(self.X_proj(x),'bt(hp)->bthp',h=self.n_heads)
        dt=nn.functional.softplus(dt)
        # Apply rotary position embedding to B and C
        cos, sin = self.BC_rotary_emb(hidden_states, position_ids=position_ids)
        C, B = apply_CB_rotary_pos_emb(C, B, cos, sin)
        try:
            y=mamba_chunk_scan_combined(x,dt,A,B,C,chunk_size=self.chunk_len,D=self.D)
        except Exception as e:
            y=ssd(x,dt,A,B,C,self.chunk_len,self.D)
        y=self.out_proj(rearrange(y,'bthp->bt(hp)').to(dtype))
        return y
    \end{lstlisting}
    \caption{Example SSD implementation in PyTorch}
\end{listing}

\clearpage

\subsection{InnerFuncAttn}
\label{sec:implementation_code:innerfuncattn}

\begin{listing}[ht]
    \begin{lstlisting}[language=Python]
def static_mask(attention_mask,q_len,kv_len):
    """
    notation: 
    attention_mask: [bsz, seq_len]
    q_len: query length
    kv_len: key and value length
    """
    bsz, seq_len = attention_mask.size()
    # Create causal mask
    causal_mask = torch.full((q_len,kv_len),float('-inf')).triu(1)
    # Expand shape
    causal_mask = causal_mask[None,None,:,:].expand(bsz,1,-1,-1)
    attention_mask = attention_mask[:,None,None,:].expand(-1,1,1,-1)
    # Apply padding
    padding = causal_mask[:,:,:,:seq_len] + attention_mask
    return causal_mask[:,:,:,:seq_len].masked_fill(padding==0,float('-inf'))
    \end{lstlisting}
    \caption{Example static attention mask method in PyTorch}
    \label{lst:static_mask}
\end{listing}

\begin{listing}[ht]
    \begin{lstlisting}[language=Python]
def dynamic_mask(attention_mask,dynamic_mask,q_len,kv_len):
    """
    notation: 
    attention_mask: [bsz, seq_len]
    dynamic_mask: [n_heads, max_position_len]
    q_len: query length
    kv_len: key and value length
    """
    bsz, seq_len = attention_mask.size()
    num_heads = dynamic_mask.size(0)
    # Create causal mask
    causal_mask = torch.full((q_len,kv_len),float('-inf')).triu(1)
    # Expand shape
    causal_mask = causal_mask[None,None,:,:].expand(bsz,num_heads,-1,-1)
    attention_mask = attention_mask[:,None,None,:].expand(-1,num_heads,1,-1)
    dynamic_mask = dynamic_mask[None,:,None,:seq_len].expand(bsz,-1,1,-1)
    # Apply padding
    padding = causal_mask[:,:,:,:seq_len] + attention_mask * dynamic_mask
    return causal_mask[:,:,:,:seq_len].masked_fill(padding==0,float('-inf'))
    \end{lstlisting}
    \caption{Example dynamic attention mask method in PyTorch}
    \label{lst:dynamic_mask}
\end{listing}

\begin{listing}[ht]
    \begin{lstlisting}[language=Python, breaklines]
class InnerFuncAttn(nn.Module):
    def __init__(self,d_model,n_heads,n_innerV,d_innerV_ret,max_position):
        super().__init__()
        self.n_heads,self.d_head=n_heads,d_model//n_heads
        # Initialize Parameters
        self.Q_proj=Linear(d_model,d_model)
        self.K_proj=Linear(d_model,d_model)
        self.dynamic_mask=Parameter(torch.ones(n_heads,max_position))
        self.V_queries=Linear(d_model,d_innerV_ret)
        self.V_keys=Parameter(torch.zeros(n_innerV,d_innerV_ret))
        self.V_embed=Embedding(n_heads,d_model)
        self.out_proj=Linear(d_model,d_model)
        # Rotary Position Embedding
        self.QK_rotary_emb=RotaryEmbedding(self.d_head,max_position)

    def inner_func(self,x):
        V_queries=self.V_queries(x)
        sim=torch.matmul(V_queries, self.V_keys.T) # einsum('btn,kn->btk')
        V_embed=self.V_embed(sim.topk(1, dim=-1).indices)
        V=x*V_embed.sum(dim=-2) # einsum('btd,btkd->btd')
        return V

    def forward(self,x,attention_mask,position_ids):
        """
        Notation: b - batch t - length d - d_model h - n_heads p - d_head
        """
        # Compute linear projection Q K and inner function V
        Q=self.Q_proj(x)
        K=self.K_proj(x)
        V=self.inner_func(x)
        # Split into multiple heads
        Q,q_len=rearrange(Q,"bt(hp)->bhtp",h=self.n_heads),Q.size(1)
        K,kv_len=rearrange(K,"bt(hp)->bhtp",h=self.n_heads),K.size(1)
        V=rearrange(V,"bt(hp)->bhtp",h=self.n_heads)
        # Apply rotary position embedding to Q and K
        cos,sin=self.QK_rotary_emb(V,position_ids=position_ids)
        Q,K=apply_QK_rotary_pos_emb(Q,K,cos,sin)
        # Compute Attention score matrix and rotary position embedding matrix
        attn_score=torch.matmul(Q,K.transpose(-2,-1))/math.sqrt(self.d_head)
        mask=dynamic_mask(attention_mask,self.dynamic_mask,q_len,kv_len)
        attn_score=attn_score+mask
        attn_score=F.softmax(attn_score,dim=-1)
        # Weighted attention score to inner function state V
        y=torch.matmul(attn_score,V)
        y=rearrange(y,"bhtp->bt(hp)")
        # Project Output
        y=self.out_proj(y)
        return y
    \end{lstlisting}
    \caption{Example InnerFuncAttn implementation in PyTorch}
    \label{lst:innerfuncattn}
\end{listing}
\clearpage

\subsection{Cross Domain Mixture of Experts}
\label{sec:implementation_code:cdmoe}

\begin{listing}[ht]
    \begin{lstlisting}[language=Python,breaklines]
class CDMoE(nn.Module):
    def __init__(self,d_model,act,d_ff,d_ret,n_experts,n_heads,k_per_head):
        super().__init__()
        self.act_fn,self.n_heads,self.k_per_head=ACT2FN[act],n_heads,k_per_head
        # Cross Domain
        self.shared_up_proj=Linear(d_model,d_ff)
        self.shared_down_proj=Linear(d_ff,d_ff_private)
        # Queries and Keys
        self.queries=Linear(d_model,d_ret*n_heads)
        self.num_keys=math.sqrt(n_experts)
        self.keys=Parameter(torch.zeros(n_heads,self.num_keys,2,d_ret//2))
        # Private Experts
        self.down_embed=Embedding(n_experts,d_model)
        self.up_embed=Embedding(n_experts,d_model)
    
    def forward(self,x):
        """
        Notation: b - batch t - length d - d_model n - d_retrieval
                  h - n_heads p - 2 for product key k - number of keys
        """
        # Compute Cross Domain
        phi_x=self.shared_down_proj(self.act_fn(self.shared_up_proj(x)))
        # Queries and Keys for Product-Key
        queries=self.queries(phi_x)
        queries=rearrange(queries,'bt(phn)->pbthn',p=2,h=self.n_heads)
        # Compute scores and indices
        sim=einsum('pbthn,hkpn->pbthk',queries,self.keys)
        (s_x,s_y),(i_x,i_y)=sim.topk(self.k_per_head,dim=-1)
        all_s=einx.add('... i,... j->... (i j)',s_x, s_y)
        all_i=einx.add('... i,... j->... (i j)',i_x*self.num_keys,i_y)
        s,pk_i=all_s.topk(self.k_per_head,dim=-1)
        i=all_i.gather(-1,pk_i)
        # Compute Private Experts
        down_embed,up_embed=self.down_embed(i),self.up_embed(i)
        experts_s=self.act_fn(einsum('btd,bthkd->bthk',phi_x,down_embed)*s)
        y=einsum('bthk,bthkd->btd',experts_s,up_embed) + phi_x
        return y
    \end{lstlisting}
    \caption{Example CDMoE implementation in PyTorch}
    \label{lst:cdmoe}
\end{listing}

\clearpage

\newpage
\section{Evaluation Parameters}
\label{sec:evaluation_parameters}

\subsection{Multi-Query Associative Recall}
\label{sec:evaluation_parameters:multi_query_associative_recall}

\begin{table}[!ht]
    \centering
    \caption{
        \textbf{Data Parameters}.
        % 我们基于原始的多查询联想召回引入更难的任务版本, 其中非query/key/value的部分被替换为随机的token. 我们还使用了更多的键值对和更长的序列长度. 对于每个序列长度 $T in \{256, 512, 1024, 2048\}$, 我们使用 $T/4$ 个键值对. 总词汇量为 $8192$, 训练样本大约为 $250k$, 测试样本为 $1k$.
        We introduce a more challenging task version based on the original multi-query associative recall~\citep{arora2024zoology}, where tokens that are not query/key/value are replaced with random tokens. We also use more key-value pairs and longer sequence lengths. For each sequence length $T \in \{256, 512, 1024, 2048\}$, we use $T/4$ key-value pairs. The total vocabulary size is $8192$, with approximately $250k$ training samples and $1k$ test samples.
    }
    \label{tab:multi_query_associative_recall:data}
    \begin{tabular}{@{}ccccccccccccccccc@{}}
    \toprule
    \sc{vocab} & \sc{seq len} & \sc{kv pairs} & \sc{train examples} & \sc{test examples} & \sc{powar a} & \sc{batch} & \sc{max epochs} \\
    \midrule
    8192 & 256 & 64 & $2^{18}$ & $2^{10}$ & 0.01 & 256 & 64 \\
    8192 & 512 & 128 & $2^{18}$ & $2^{10}$ & 0.01 & 128 & 64 \\
    8192 & 1024 & 256 & $2^{18}$ & $2^{10}$ & 0.01 & 64 & 64 \\
    8192 & 2048 & 512 & $2^{18}$ & $2^{10}$ & 0.01 & 32 & 64 \\
    \bottomrule
    \end{tabular}
\end{table}

\begin{table}[!ht]
    \centering
    \caption{
        \textbf{Model Parameters}.
        % 这些算法都可以分割多头, 所以我们设置为单头, 并使用常见的单头维度 $d_{model} \in \{32, 64, 128, 256\}$. 为了公平起见, SSD 算法与 Mamba2 中的验证结构有所不同, 我们移除了一维因果卷积与门控MLP. 所有算法统一使用序列变换到状态变换的结构并堆叠两层. 为后续算法混合做准备, 这些算法每个维度的学习率都相同.
        These algorithms can all split into multiple heads, so we set them to single heads and use common single head dimensions $d_{model} \in \{32, 64, 128, 256\}$. For fairness, the SSD algorithm is different from the validation structure in Mamba2~\citep{mamba2} and we remove the one-dimensional causal convolution and gated MLP. All algorithms use the structure of sequence transformation to state transformation and stack 2 layers. In preparation for subsequent algorithm mixing, the learning rate for each dimension of these algorithms is the same.
    }
    \label{tab:multi_query_associative_recall:model}
    \begin{tabular}{@{}ccccccccccc@{}}
    \toprule
    \sc{Algorithm} & $d_{model}$ & $n_{layers}$ & $n_{heads}$ & $d_{state}$ & chunk\_len & $n_{v}$ & $d_{innerV}$ & \sc{leaning rate}\\
    \midrule
    QCAttn & 32/64/128/256 & 2 & 1 & --- & --- & --- & --- & 4e-4/3e-4/2e-4/1e-4 \\
    SSD & 32/64/128/256 & 2 & 1 & 128 & 256 & --- & --- & 4e-4/3e-4/2e-4/1e-4 \\
    InnerFuncAttn & 32/64/128/256 & 2 & 1 & --- & --- & 2 & 16/32/64/128 & 4e-4/3e-4/2e-4/1e-4 \\
    \bottomrule
    \end{tabular}
\end{table}

\begin{figure}[!ht]
    \centering
    \includegraphics[width=\textwidth]{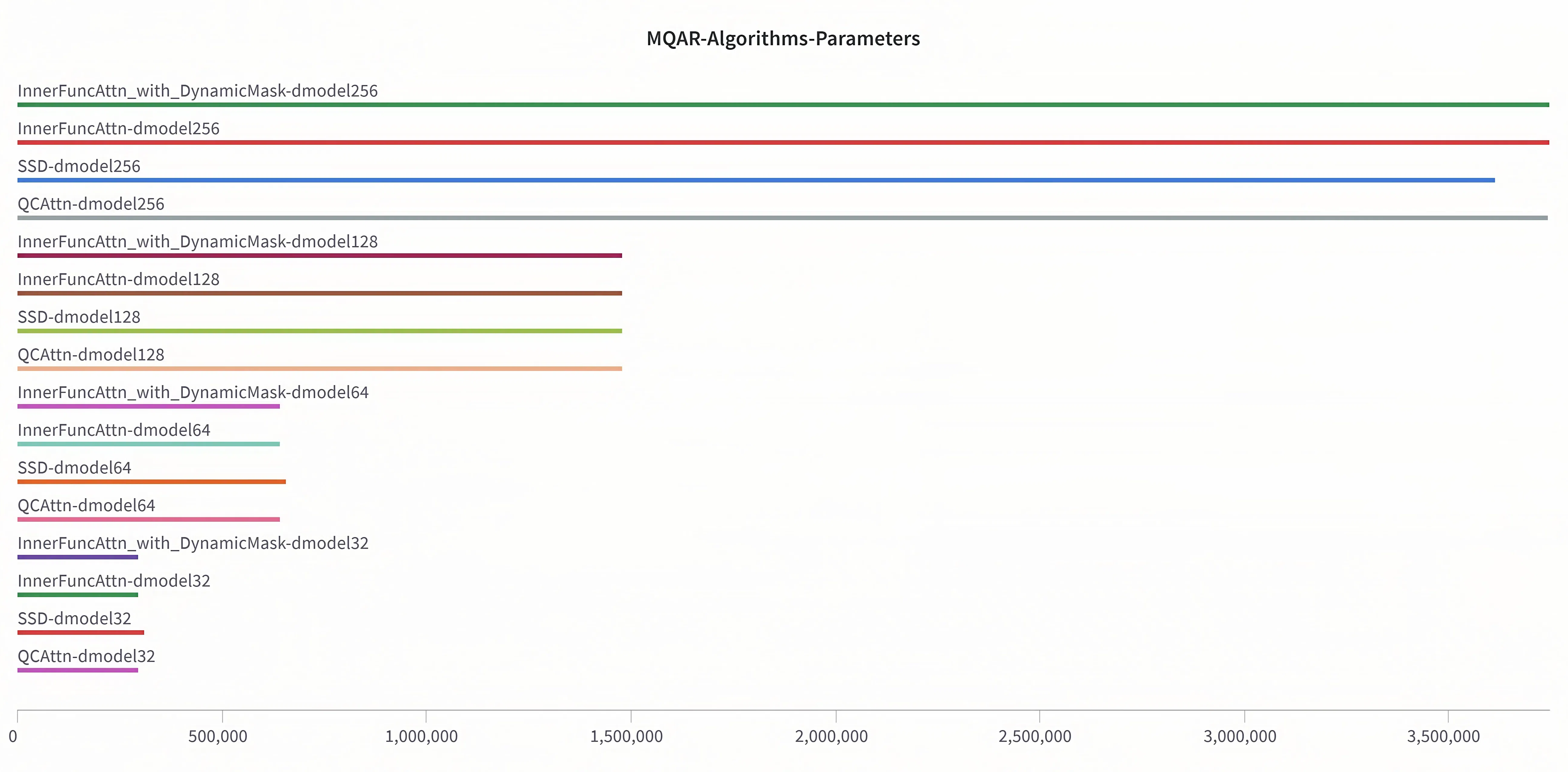}
    \caption{
        \textbf{Different Algorithms Parameters}.
        % 内函数注意力无论是否添加动态注意力掩码, 在不同维度规模下, 参数量都与 QCAttn 相差不大. SSD 在增加维度规模时, 参数量增加较少.
        Whether adding dynamic attention mask or not, the number of parameters of InnerFuncAttn is similar to QCAttn at different dimensional scales. SSD increases the number of parameters less when increasing the dimensional scale.
    }
\end{figure}

\subsection{Downstream Evaluation}
\label{sec:evaluation_parameters:downstream_evaluation}

% 为了避免训练数据不同导致在下游任务的分数偏差, 我们重新训练了四个模型架构, 包括使用 QCAttn 算法的 Llama, 使用 SSD 算法的 Mamba2, 使用混合算法的 Jamba, 以及我们的架构. 所有模型都在Smollm-Corpus数据集上完成训练, 使用 NeoX 分词器. 训练360M和1.3B两个规模的模型, 参数参考表. 训练环境是Nvidia开源的PyTorch镜像24.2版本, 此版本兼容mamba-ssm库中的cuda内核SSD算法. 使用Transformers库的Trainer类完成训练. 使用AdamW优化器, $\beta_1=0.9, \beta_2=0.999$ 和 $weight\_decay=0.01$. 线性热身步骤为总步数的$10\%$, 达到最大学习率$2e-4$, 然后余弦衰减到最小学习率$2e-5$. 无偏置项. RMSNorm替换LayerNorm.

To avoid score bias in downstream tasks due to different training data, we retrain four model architectures, including Llama using the QCAttn algorithm, Mamba2 using the SSD algorithm, Jamba using the hybrid of QCAttn and SSD, and our architecture. We train models of two scales, 360M and 1.3B, with parameters referenced in the table\ref{tab:downstream_evaluation:model}.

\begin{itemize}
    \item All models are trained on the Smollm-Corpus~\citep{benallal2024smollmcorpus} dataset using the NeoX tokenizer.
    \item The training environment is the Nvidia open-source PyTorch image~\citep{pytorch} version 24.2, which is compatible with the cuda kernel SSD algorithm in the mamba-ssm library.
    \item Training is completed using the Trainer class in the Transformers~\citep{wolf-etal-2020-transformers} library.
    \item AdamW optimizer hyperparameters $\beta_1=0.9, \beta_2=0.999$ and $weight\_decay=0.01$.
    \item The linear warm-up steps are $10\%$ of the total steps, reaching the maximum learning rate of $2e-4$, and then cosine decay to the minimum learning rate of $2e-5$.
    \item No bias terms.
    \item RMSNorm instead of LayerNorm.
\end{itemize}

% 为了下游评估, 我们使用Eleuther的 LM evaluation harness, 验证集包括下面这些任务: LAMBADA, MMLU, TriviaQA, ARC, PIQA, HellaSwag, OBQA, Winogrande
For downstream evaluation, we use LM evaluation harness from EleutherAI~\citep{eval-harness}, the validation dataset includes the following tasks:

\begin{itemize}
    \item MMLU~\citep{hendrycks2021measuringmassivemultitasklanguage}
    \item TriviaQA~\citep{m2017triviaqa}
    \item ARC~\citep{clark2018think}
    \item PIQA~\citep{bisk2020piqa}
    \item HellaSwag~\citep{zellers2019hellaswag}
    \item OBQA~\citep{mihaylov2018can}
    \item Winogrande~\citep{sakaguchi2021winogrande}
\end{itemize}

\begin{table}[!ht]
    \centering
    \caption{
        \textbf{Model Parameters}.
        % 为了公平起见, 我们尽可能的将这四个模型的重要参数调整为相同大小, 并通过在LlaMa与Mamba2中加入路由混合专家, 与仔细调整LlaMa, Mamba2和Jamba的前馈网络扩展大小来保证四个模型的总参数与激活参数尽可能接近, 最终我们得到320M与1.3B两个规模的模型.
        For fairness, we adjust the important parameters of these four models to be as close in size as possible, and ensure that the total parameters and activation parameters of the four models are as close as possible by adding routing mixture of experts in LlaMa and Mamba2, and carefully adjusting the feedforward network expansion size of LlaMa, Mamba2, and Jamba. Finally, we obtain models of two scales, 320M and 1.3B.
    }
    \label{tab:downstream_evaluation:model}
    \begin{tabular}{@{}ccccccccccc@{}}
    \toprule
    \sc{Model} & $d_{model}$ & $n_{layers}$ & $n_{heads}$ & $d_{state}$ & chunk\_len & $n_{v}$ & $n_{experts}$ & \sc{leaning rate} & \sc{batch size} \\
    \midrule
    LlaMa-320M & 768 & 24 & 12 & --- & --- & --- & 4 & 3e-4 & 1M tokens \\
    Mamba2-320M & 768 & 24 & 12 & 128 & 256 & --- & 4 & 3e-4 & 1M tokens \\
    Jamba-320M & 768 & 24 & 12 & 128 & 256 & --- & 4 & 3e-4 & 1M tokens \\
    Cheems-320M & 768 & 24 & 12 & 128 & 256 & 6 & 3072 & 3e-4 & 1M tokens \\
    \midrule
    LlaMa-1.3B & 2048 & 24 & 32 & --- & --- & --- & 4 & 2e-4 & 2M tokens \\
    Mamba2-1.3B & 2048 & 24 & 32 & 128 & 256 & --- & 4 & 2e-4 & 2M tokens \\
    Jamba-1.3B & 2048 & 24 & 32 & 128 & 256 & --- & 4 & 2e-4 & 2M tokens \\
    Cheems-1.3B & 2048 & 24 & 32 & 128 & 256 & 16 & 8192 & 2e-4 & 2M tokens \\
    \bottomrule
    \end{tabular}
\end{table}
  
\end{document}